\documentclass[runningheads]{llncs}
\usepackage[T1]{fontenc}
\usepackage{graphicx}
\usepackage{hyperref}
\usepackage{xcolor}

\usepackage[acronym]{glossaries}
\usepackage{microtype}
\usepackage{float}
\usepackage{subfigure}
\usepackage{booktabs} 
\usepackage{wrapfig}
\usepackage[utf8]{inputenc}

\usepackage[capitalize,noabbrev]{cleveref}

\newacronym{mpc}{MPC}{Model Predictive Control}
\newacronym{gnn}{GNN}{Graph Neural Network}
\newacronym{cnn}{CNN}{Convolutional Neural Network}
\newacronym{mse}{MSE}{Mean Squared Error}
\newacronym{mae}{MAE}{Mean Absolute Error}
\newacronym{lpips}{LPIPS}{Learned Perceptual Image Patch Similarity}
\newacronym{ssim}{SSIM}{Structural Similarity}
\newacronym{psnr}{PSNR}{Peak Signal-to-Noise Ratio}
\newacronym{vqa}{VQA}{Visual Question Answering}
\newacronym{mlp}{MLP}{Multi Layer Perceptron}
\newacronym{vae}{VAE}{Variational Autoencoder}
\newacronym{pcg}{PCG}{Procedural Content Generation}
\newacronym{savi}{SAVi}{Slot Attention for Video}
\newacronym{slam}{SLAM}{Simultaneous Localization And Mapping}
\newacronym{paig}{PAIG}{Physics-As-Inverse-Graphics}
\newacronym{lstm}{LSTM}{Long Short-Term Memory}
\newacronym{pde}{PDE}{Partial Differential Equation}

\begin{document}
\title{ViPro: Enabling and Controlling Video Prediction for Complex Dynamical Scenarios using Procedural Knowledge}
\titlerunning{Enabling Video Prediction using Procedural Knowledge}
%

\author{Patrick Takenaka\inst{1,2}\orcidID{0009-0005-2295-562X} \and
Johannes Maucher\inst{1}\orcidID{0000-0002-3804-8937} \and
Marco F. Huber\inst{2,3}\orcidID{0000-0002-8250-2092}
{\tt\small \{takenaka,maucher\}@hdm-stuttgart.de, marco.huber@ieee.org}
}

\authorrunning{Takenaka et al.}
%
\institute{Institute for Applied AI, Hochschule der Medien Stuttgart, Germany\and
Institute of Industrial Manufacturing and Management IFF, University of Stuttgart, Germany\and
Fraunhofer Institute for Manufacturing Engineering and Automation IPA, Stuttgart, Germany \\
}

\maketitle              
\begin{abstract}
We propose a novel architecture design for video prediction in order to utilize procedural domain knowledge directly as part of the computational graph of data-driven models. On the basis of new challenging scenarios we show that state-of-the-art video predictors struggle in complex dynamical settings, and highlight that the introduction of prior process knowledge makes their learning problem feasible. Our approach results in the learning of a symbolically addressable interface between data-driven aspects in the model and our dedicated procedural knowledge module, which we utilize in downstream control tasks. 

\keywords{Informed Machine Learning \and Procedural Knowledge \and Video Prediction.}
\end{abstract}
\section{Introduction}
Current advances in deep learning research~\cite{brownLanguageModelsAre2020,jumperHighlyAccurateProtein2021,ouyangTrainingLanguageModels2022} have shown the tremendous potential of data-driven approaches. However, upon taking a closer look, oftentimes domain-specific inductive biases in the training process or the architecture~\cite{jumperHighlyAccurateProtein2021} play a critical role in making the most of the available data. While an end-to-end generalized architecture that is able to tackle a wide range of problems is elegant, we argue that this often leads to models that require large amounts of data to be feasible while still being unable to generalize well outside of the training distribution~\cite{marcusRebootingAIBuilding2019}, leading to limited applicability for more specialized use cases where data are often scarce, such as, for instance, in the medical domain. Transfer of deep learning research into applications usually requires further fine-tuning of the model for the given use case, and this often corresponds to collecting specialized data. We believe that providing additional means for domain experts---who may or may not be experts in machine learning---to represent their knowledge besides data in deep learning models is crucial for driving AI adaption in more areas. Ideally, specializing the model to a domain should reduce the complexity of the learning task and thus, lead to leaner architectures that require less data than a domain agnostic variant, while providing better predictions for sparsely observed situations. Furthermore, as we show with our approach, such grey box modelling approaches also inherently increase the controllability of the model. Thus, developing a model that benefits from both domain knowledge and data samples together is our objective. 

There are various types of domain knowledge that can be integrated, and ways how they can be integrated~\cite{vonruedenInformedMachineLearning2023}, ranging from using logic rules or differential equations to structure and expand both the loss function and the learning process~\cite{xuSemanticLossFunction2018,raissiPhysicsinformedNeuralNetworks2019}, to architectural considerations that take into account the structure of the underlying problem, such as \glspl*{gnn} \cite{wattersVisualInteractionNetworks2017a} to model interactions, or more famously to use \glspl*{cnn} for spatial data. 

We propose to view the knowledge integration types from a different perspective and group them in either procedural or declarative knowledge. While the latter encompasses domain facts or rules (``Knowing-That''), procedural knowledge represents a more abstract level of information by describing a domain process irrespective of any concrete observation (``Knowing-How''). A straightforward representative of this paradigm would be a mathematical formula, which can describe a certain relation between variables in a concise manner without risk of running into a long-tail problem for less frequently observed variable assignments as would be the case in data-driven approaches. Without a doubt, however, data-driven function approximators are immensely successful in the real world because we are often not able---or it is impossible---to describe the problem in such a precise way. However, we argue that currently data-driven approaches are often used without considering whether the underlying mechanisms of the domain could be described in a more efficient manner without resorting to arduous data collection.

In this work, we propose a novel architecture design that integrates such procedural knowledge as an independent module into the overall architecture. We apply it to video prediction, a task where state-of-the-art models often still struggle due to the high spatio-temporal complexity involved in scenes. Its environments often involve understanding complex domain processes that are hard to robustly learn from observations only and thus are likely to benefit from domain inductive biases. At the same time, this field is the foundation for many possible downstream tasks such as \gls*{vqa}~\cite{wuSlotFormerUnsupervisedVisual2023}, \gls*{mpc}~\cite{jaquesPhysicsasInverseGraphicsUnsupervisedPhysical2019}, or system identification \cite{jaquesPhysicsasInverseGraphicsUnsupervisedPhysical2019}. We create several scenarios which feature complex dynamics, and integrate the knowledge about these dynamics. We show that current deep learning models struggle on their own, but can thrive once enhanced with procedural domain knowledge. We verify that this is still possible even with very limited data, and further highlight that such an interface enables control in the target domain w.r.t. the integrated function parameters at test time, providing a potential basis for downstream control tasks and allowing flexibility in adjusting the model for novel scene dynamics.

In summary, our contributions are:
\begin{itemize}
    \item Specification and analysis of an architectural design for interfacing procedural knowledge with a data-driven model.
    \item Introduction of novel challenging scenarios with complex dynamics for video prediction.
    \item Application to a downstream control task by relying on the inherently achieved disentanglement w.r.t. the function parameters.
\end{itemize}

The paper is structured as follows: First, relevant related work is shown in \cref{sec:related}, after which our proposed procedural knowledge integration scheme is introduced in \cref{sec:arch}, followed by a description of our datasets in \cref{sec:dataset}. In \cref{sec:exp}, we first describe our used setup including implementation details in \cref{sec:setup} and continue by establishing baseline results in \cref{sec:baseline}. We then analyze the latent state of the model and the resulting controllability in \cref{sec:control}. We conclude by discussing limitations and potential directions for future work in \cref{sec:limits}. Our datasets and code are available at \url{https://github.com/P-Takenaka/nesy2024-vipro}.

\section{Related Work}\label{sec:related}
Predicting future video frames is a challenging task that requires certain inductive biases in the training process or model architecture in order to lead to acceptable prediction outcomes. The most commonly integrated bias is the modelling of the temporal dependency between individual frames, which assumes that future frames are dependent on past frames~\cite{wangPredRNNRecurrentNeural2017,dentonStochasticVideoGeneration2018,wangPredRNNRecurrentNeural2023}. Some methods also exploit the fact that the scene is composed of objects by structuring the latent space accordingly~\cite{kosiorekSequentialAttendInfer2018a,linImprovingGenerativeImagination2020,wuSlotFormerUnsupervisedVisual2023,traubLearningWhatWhere2023}, which improved scene reconstruction performance further compared to approaches that rely on a single global latent scene representation for predictions. 

Since many dynamics in video scenes are of a physical nature, there are also works that explore the learning of the underlying \glspl*{pde} to facilitate video reconstruction~\cite{leguenDisentanglingPhysicalDynamics2020,donaPDEDrivenSpatiotemporalDisentanglement2021,yangLearningPhysicsConstrained2022,wuDisentanglingStochasticPDE2023}. Another line of work---most similar to our approach---considers a more explicit representation of dynamics knowledge in the model~\cite{jaquesPhysicsasInverseGraphicsUnsupervisedPhysical2019,kandukuriPhysicalRepresentationLearning2022}. Here, discretized \glspl*{pde} are integrated to calculate a physical state for each frame, which is decoded back into an image representation. These approaches were, however, limited to 2D dynamics of sprites, which allowed the dynamics model to operate directly in the screen space, making the learning problem much easier, while limiting applicability to more realistic settings. These methods also relied on the Spatial Transformer Network~\cite{NIPS2015_33ceb07b} for decoding purposes. Dynamical properties of the scene besides the object positions---such as for instance changing lighting conditions or object orientations---are not modelled with this approach, since it is based on sampling pixel predictions directly from the input reference frames. More recently, an architecture was proposed in a preliminary workshop publication~\cite{takenakaGuidingVideoPrediction2023} that could in theory handle such dynamic properties. It was, however, only applied to semantic segmentation in visually simple settings and we show that it fails at video reconstruction in our datasets. Our work continues in this line of research and increases the complexity of video scenarios that can be handled, while broadening the applicability by bridging the gap to control-based downstream tasks such as \gls*{mpc}.

\section{Proposed Architecture}\label{sec:arch}
Our objective is to allow domain experts to integrate their knowledge of underlying domain processes in a data-driven architecture in a domain-independent way. As such, we embed this knowledge represented as a \emph{programmatic function} $F$ within a distinct \emph{procedural knowledge module} $P$ inside the overall architecture. Instead of learning the domain dynamics itself, we thus provide the means for the model to learn the interface between $F$ and its data-driven components. This is possible since we can directly execute the program code that is $F$---as opposed to for instance natural language instructions that would need some kind of encoding first---and make it part of the computational graph.

We opt for an auto-regressive frame prediction scheme in which the model is exposed to the initial $n$ video frames in order to learn the data sample dynamics, before it auto-regressively predicts the next $m$ frames on its own. Learning is guided by the reconstruction loss 
\begin{equation}
    \mathcal{L}_{\mathrm{rec}} = \frac{1}{N}\sum_{i=0}^N (\hat{V}_i - V_i)^2
\end{equation}
of all $N = m + n$ predicted RGB frames $\hat{V}$ and the ground-truth $V$.

\begin{figure}[t]
\begin{center}
    \includegraphics[width=.9\linewidth]{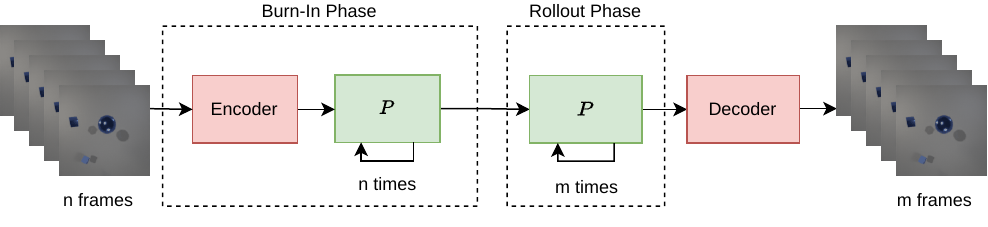}
    \caption{Overview of our auto-regressive video prediction process. The first $n$ frames are used as reference and are encoded by the model in order to obtain an initial latent representation of the scene. After this burn-in phase the model has to rollout future $m$ frames on its own.}
    \label{fig:rollout_overview}
    \end{center}
\end{figure}

Our architecture is thus composed of three main components: 1) An initial video frame encoder, which embeds the $n$ observed frames into a suitable latent representation, 2) our \emph{procedural knowledge module} $P$ that transforms the frame's latent representation to the next time step, and 3) a final video frame decoder that transforms the latent representations back into an image representation, as depicted in \cref{fig:rollout_overview}.

\begin{figure}[t]
\begin{center}
\begin{minipage}{.65\textwidth}
    \includegraphics[width=\linewidth]{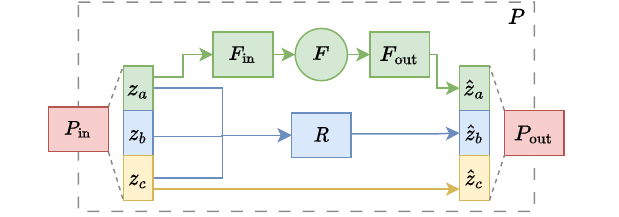}
\end{minipage}
\begin{minipage}{.3\textwidth}
    \includegraphics[width=\linewidth]{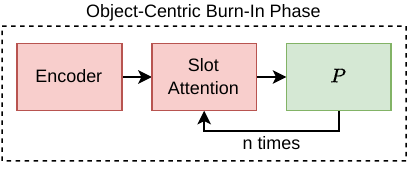}
\end{minipage}
\caption{\textbf{Left:} Structure of our proposed \emph{procedural knowledge module} $P$. \textbf{Right:} Abstract view of the burn-in phase for the object-centric variant of our architecture.}
    \label{fig:abstract_integration}
\end{center}
\end{figure}

The core of our contribution resides in $P$ (see \cref{fig:abstract_integration}): It combines the integrated \emph{programmatic function} $F$ with a deep \emph{spatio-temporal prediction model} $R$ to obtain the latent frame representation $\hat{z}$ of the next time step based on the latent representation $z$ of the current step. We implement this by transforming the latent vector $z$ into a representation that is separable into three components $z_a$, $z_b$, and $z_c$ via a latent encoder $P_{\mathrm{in}}$ and a decoder $P_{\mathrm{out}}$. Each part $a,b,c$ is responsible for encoding different aspects of the frame that are learned implicitly through our architectural design, which we now describe and also later verify in our experiments.

$z_a$ represents features that are relevant for $F$---i.e. which correspond to its input and output parameters---in latent space. Since $F$ is a programmatic function, we can directly integrate it into the computational graph as an individual layer. However, since its input space is usually symbolic and not in a latent representation, $z_a$ is translated into and from its symbolic pendant $s$ through dedicated mapping layers $F_\mathrm{in}$ and $F_\mathrm{out}$, respectively. We further facilitate the transformation between latent and symbolic representation of the state by introducing a regularization loss $\mathcal{L}_s$ that is based on the symbolic state~$s$ and its auto-encoded version for all $N$ predicted frames:

\begin{equation}
    \mathcal{L}_s = \frac{1}{N}\sum_{i=0}^{N} (s_i - F_{\mathrm{in}}(F_{\mathrm{out}}(s_i)))^2
\end{equation}

The final loss is thus $\mathcal{L} = \mathcal{L}_{\mathrm{rec}} + \lambda\mathcal{L}_s$, with $\lambda$ being a constant weighting coefficient.

The second part $z_b$ contains residual scene dynamics that are not covered by $F$. It corresponds to the output of $R$, which is a typical auto-regressive frame predictor model~\cite{wuSlotFormerUnsupervisedVisual2023} with the goal of modelling spatio-temporal scene properties. It takes into account the whole latent representation consisting of $z_a$, $z_b$, and $z_c$. Instead of predicting the full future latent state on its own, we show that when combined with $F$, this model $R$ focuses on residual scene dynamics only and does not interfere with dynamics handled by $F$. This property allows us to exert a certain level of control when predicting frames w.r.t. the inputs and outputs of $F$, as shown in \cref{sec:control}.

Finally, $z_c$ contains static scene features that do not change frame-to-frame, such as for instance parts of the background or object colors. It is thus realized as a residual connection directly mapping from $z_c$ to $\hat{z}_c$. Auto-regressive approaches often suffer from accumulating errors, as future predictions build on top of past prediction mistakes. We were able to alleviate this issue by allowing the model to store scene statics in $z_c$.

Thus, our model has three different pathways available to pass latent features from one video time step to the next. Each contains meaningful inductive biases that encourage the model to follow our interpretation of the latent vectors. $z_a$ contains only the encoded state required for $F$ and is limited by an information bottleneck induced by the symbolic state transformation for $F$. $z_c$ is a direct shortcut from the first encoded frame, and thus offers an easy to learn path for time-invariant features. Finally, $z_b$ offers the only unrestricted pathway that allows encoding time-variant features. Keeping representations intact over many frames in auto-regressive models such as $R$ is a challenging learning objective however, which 1) makes it less likely that the model will try to encode static scene features that are better handled by $z_c$, and 2) encourages the model to rely on $z_a$ for dynamics that are handled by $F$.

Object-centric representations have shown to further improve video prediction performance over methods that work with a single latent state for the whole scene~\cite{wuSlotFormerUnsupervisedVisual2023}. Since some of our datasets are object based, we also introduce a variant of our proposed architecture that operates on slot-based object representations. For our module $P$ the main difference is that the latent state of a single frame is now separated into multiple object representations. In practice, this amounts to an additional dimension in the tensors which can be handled in parallel by all components of $P$. During the burn-in phase, our module $P$ takes turns with a slot attention~\cite{locatelloObjectCentricLearningSlot2020} module to assemble these frame object representations. For each frame, the latter refines slot-based object representations w.r.t. the encoded video frame of the current time step and the object representation prediction of our module $P$ from the last time step (cf. \cref{fig:abstract_integration}). Afterwards, $P$ uses this refined representation to predict the next time step.

With our architecture we aim to concentrate the domain inductive biases within $P$, and keep the image encoder and decoder parts domain agnostic and in line with current state-of-the-art models in video prediction. As opposed to models that use domain-specific decoders such as neural renderers~\cite{murthyGradSimDifferentiableSimulation2020}, we thus do not need to obtain a full symbolic representation of the scene and can keep parts not relevant for $F$ in the latent space of the model. At the same time, the architecture's modularity facilitates the integration of and interfacing with future developments in video prediction, such as better image encoder or decoders but also better predictors in place of $R$.

\section{Data}\label{sec:dataset}
We designed and introduce three new datasets which consist of videos of rendered 3D scenes using Kubric \cite{greffKubricScalableDataset2022}. They feature complex, nonlinear physical dynamics and aim to provide suitable testing grounds for our approach. Other video dynamics datasets in the literature often either happened in 2D screen space only ~\cite{jaquesPhysicsasInverseGraphicsUnsupervisedPhysical2019,donaPDEDrivenSpatiotemporalDisentanglement2021}, or involved only short-term dynamics for which linear approximations of the underlying dynamics were sufficient~\cite{wuSlotFormerUnsupervisedVisual2023,kipfConditionalObjectCentricLearning2022}. As we also show in our experiments, such models fail once the underlying dynamics increase in complexity as in our datasets. Thus, our settings are an ideal candidate for which the introduction of domain knowledge can supplement---or even enable---purely data-driven approaches. We integrate the dynamics equations of each dataset as $F$ within the \emph{procedural knowledge module} $P$ of our model.

Our main and most challenging dataset ``Orbits'' is based on the three body problem~\cite{musielakThreebodyProblem2014}, in which multiple objects attract each other, resulting in chaotic movements once three or more objects are involved. A similar dataset was introduced in previous related work~\cite{takenakaGuidingVideoPrediction2023}, albeit with no background and only sphere object shapes.

Our second dataset is a rendering of an Acrobot environment, a setting which is commonly used for benchmarking physics and control models~\cite{brockmanOpenAIGym2016}. It features a double pendulum where one end is fixed in space, resulting in dynamics that are inherently complex with chaotic movements without further actuation. As we show in our experiments these environments are easier to predict in pixel-space for data-driven models compared to our Orbits dataset, however stabilizing the pole by actuating the joint between the pendulums is a common control objective that we want to work with as a downstream task.

Our third dataset features a non object-centric variant of the Acrobot setting. Instead of observing the double pendulum directly we mount the camera on it, while orienting it towards a static, texturized background, whose texture is the same for all data samples. We let the double pendulum move according to the dynamics defined in the Acrobot setting, and the now also moving camera observes different parts of the background texture in each frame. The prediction of future frames is thus only successful if the model is able to correctly estimate and utlize the indirectly observed pendulum dynamics. The overall dataset setup is otherwise the same as in the Acrobot setting. This dataset is related to visual \gls*{slam}, in which models establish a map of the environment based on video input, since an internal representation of the whole---always only partially observed---scene is necessary. By applying our approach to this dataset we also show its potential for visual robotics navigation tasks as future work, for which we could exchange the pendulum dynamics for a robot dynamics model in $F$.

 We use our object-centric architecture variant for the Orbits and Acrobot dataset, and the non object-centric variant for the Pendulum Camera dataset. We describe further details of these datasets in \cref{apd:dataset} and their underlying dynamics equations in \cref{apd:f}.

\section{Experiments}\label{sec:exp}
In the following we first describe our experimental setup and continue by analyzing our proposed approach w.r.t. its performance in contrast to existing methods. Afterwards we study the feasibility of using our model for downstream control tasks. 

\subsection{Setup}\label{sec:setup}
All models observe the initial six video frames and---where applicable---the symbolic input for $F$ for the very first frame. We evaluate the performance based on the prediction performance of the next 24 frames, however during training only the next twelve frames contribute to the loss in order to observe generalization performance for more rollouts.

We compare the performance with two groups of relevant state-of-the-art work in video prediction: 1) Purely data-driven approaches that do not rely on physical inductive biases in the architecture such as Slot Diffusion~\cite{wuSlotDiffusionObjectCentricGenerative2023a}, SlotFormer~\cite{wuSlotFormerUnsupervisedVisual2023} and PredRNN-V2~\cite{wangPredRNNRecurrentNeural2023} and 2) approaches that include general physical inductive biases such as PhyDNet~\cite{leguenDisentanglingPhysicalDynamics2020}, Dona et  al.~\cite{donaPDEDrivenSpatiotemporalDisentanglement2021}, and Takenaka et al.~\cite{takenakaGuidingVideoPrediction2023}. We describe the details of the configurations of these models in \cref{apd:relatedwork}.

We measure the averaged reconstruction performance for three different random seeds with the \gls*{lpips}, which has shown better alignment with human perception than other metrics by relying on a pretrained image encoder. However, for completeness we also report the standard metrics \gls*{ssim} and \gls*{psnr}. 

\subsubsection{Implementation Details}
Our used video frame encoder is a standard \gls*{cnn} with a subsequent position embedding. The modules $P_\mathrm{in}$ and $P_\mathrm{out}$ within $P$ are implemented as fully-connected networks including a single hidden layer with the ReLU activation function, whose non-linearity enables the model to learn the separable latent space that we require without relying on the learning capacity of the video frame encoder. We linearly transform $z_a$ from the symbolic to the latent space and back by implementing $F_\mathrm{in}$ and $F_\mathrm{out}$ as fully-connected layers without bias neurons. This enables easy auto-encoding between both spaces while within $P$, and forces the model to learn the complex transformation between the symbolic state for $F$ and its latent version that can be used by the video frame decoder in $P_\mathrm{in}$ and $P_\mathrm{out}$. The \emph{spatio-temporal prediction model} $R$ within $P$ is implemented as a transformer~\cite{vaswaniAttentionAllYou2017} with temporal position embedding. Finally, we use a Spatial Broadcast Decoder~\cite{wattersSpatialBroadcastDecoder2019b} as our video frame decoder.

For the object-centric variant of our model we use Slot Attention~\cite{locatelloObjectCentricLearningSlot2020} to obtain object representations in the latent space. For a scene with $M$ objects and $N$ rollout frames, we thus obtain $N\times M$ latent representations, which are decoded individually into the image space by the video frame decoder. The final frame prediction is assembled by combining the $M$ object representations as in \gls*{savi}~\cite{kipfConditionalObjectCentricLearning2022}, which predicts an auxiliary masking channel in addition to the RGB channels to weigh the individual object frame decodings.

More implementation details can be found in \cref{apd:arch}, and details about the integrated functions in \cref{apd:f}.

\subsection{Video Prediction}\label{sec:baseline}
Here we establish and discuss the video prediction performance of our proposed architecture and compare it with related work. More qualitative results for all datasets can be found in \cref{apd:vis}.

\subsubsection{Orbits}

\begin{table}[t]
    \centering
    \caption{Performance comparison of our proposed complete architecture (Base) with ablations and relevant related work for the Orbits dataset.}
    \begin{tabular}{p{0.2cm}p{4.0cm}|p{1.3cm}|p{1.3cm}|p{1.3cm}}
\toprule
                      & & \textbf{LPIPS}$\downarrow$ & \textbf{SSIM}$\uparrow$ & \textbf{PSNR}$\uparrow$ \\\midrule
                    \textbf{Ours}& & $\mathbf{4.0}$\textcolor{darkgray}{\scriptsize$\pm1$} &  $\mathbf{97.2}$\textcolor{darkgray}{\scriptsize$\pm0$} &  $\mathbf{34.7}$\textcolor{darkgray}{\scriptsize$\pm0$}\\
                    &Limited Data &  $19.5$\textcolor{darkgray}{\scriptsize$\pm2$} &  $87.1$\textcolor{darkgray}{\scriptsize$\pm0$} &  $27.7$\textcolor{darkgray}{\scriptsize$\pm0$} \\
                    &Learned Parameters in $F$ &  $\mathbf{3.9}$\textcolor{darkgray}{\scriptsize$\pm1$} &  $\mathbf{97.1}$\textcolor{darkgray}{\scriptsize$\pm0$} &  $\mathbf{34.7}$\textcolor{darkgray}{\scriptsize$\pm1$} \\
                    
                    \midrule
                    \multicolumn{2}{l|}{\textbf{Architecture Ablations}} &&& \\
                    &1) No $z_c$ & $5.2$\textcolor{darkgray}{\scriptsize$\pm0$} &  $96.4$\textcolor{darkgray}{\scriptsize$\pm0$} &  $33.8$\textcolor{darkgray}{\scriptsize$\pm0$} \\
                    &2) No $z_b, R$ &$5.7$\textcolor{darkgray}{\scriptsize$\pm0$} &  $95.8$\textcolor{darkgray}{\scriptsize$\pm0$} &  $33.0$\textcolor{darkgray}{\scriptsize$\pm0$} \\

                    &3) $F$ as Identity         &  $35.5$\textcolor{darkgray}{\scriptsize$\pm5$} &  $77.9$\textcolor{darkgray}{\scriptsize$\pm1$} &  $24.2$\textcolor{darkgray}{\scriptsize$\pm1$} \\
                    &4) Only $R$ instead of $P$   &   $41.8$\textcolor{darkgray}{\scriptsize$\pm0$} &   $76.8$\textcolor{darkgray}{\scriptsize$\pm0$} &  $23.5$\textcolor{darkgray}{\scriptsize$\pm0$}\\
                    
                    \midrule
                    \multicolumn{2}{l|}{\textbf{Related Work}} &&& \\
                    &Slot Diffusion & $26.7$\textcolor{darkgray}{\scriptsize$\pm1$} & $68.7$\textcolor{darkgray}{\scriptsize$\pm1$} & $21.9$\textcolor{darkgray}{\scriptsize$\pm0$}   \\
                    &Takenaka et al. &  $36.8$\textcolor{darkgray}{\scriptsize$\pm0$} &  $78.1$\textcolor{darkgray}{\scriptsize$\pm0$} &  $24.8$\textcolor{darkgray}{\scriptsize$\pm0$}  \\
                    &Slotformer &   $34.2$\textcolor{darkgray}{\scriptsize$\pm0$} &  $75.3$\textcolor{darkgray}{\scriptsize$\pm0$} &  $23.2$\textcolor{darkgray}{\scriptsize$\pm0$} \\
                    &PhyDNet    &  $35.7$\textcolor{darkgray}{\scriptsize$\pm1$} &  $77.6$\textcolor{darkgray}{\scriptsize$\pm0$} &  $24.0$\textcolor{darkgray}{\scriptsize$\pm0$} \\
                    &PredRNN-V2 & $34.4$\textcolor{darkgray}{\scriptsize$\pm0$} &  $78.5$\textcolor{darkgray}{\scriptsize$\pm0$} &  $24.3$\textcolor{darkgray}{\scriptsize$\pm0$} \\
                    &Dona et al. & $41.1$\textcolor{darkgray}{\scriptsize$\pm0$} &  $76.7$\textcolor{darkgray}{\scriptsize$\pm0$} &  $23.5$\textcolor{darkgray}{\scriptsize$\pm0$} \\
\bottomrule
\end{tabular}
    \label{tab:base_performance}
\end{table}

\begin{figure}[t]
    \centering
    \includegraphics[width=.8\linewidth]{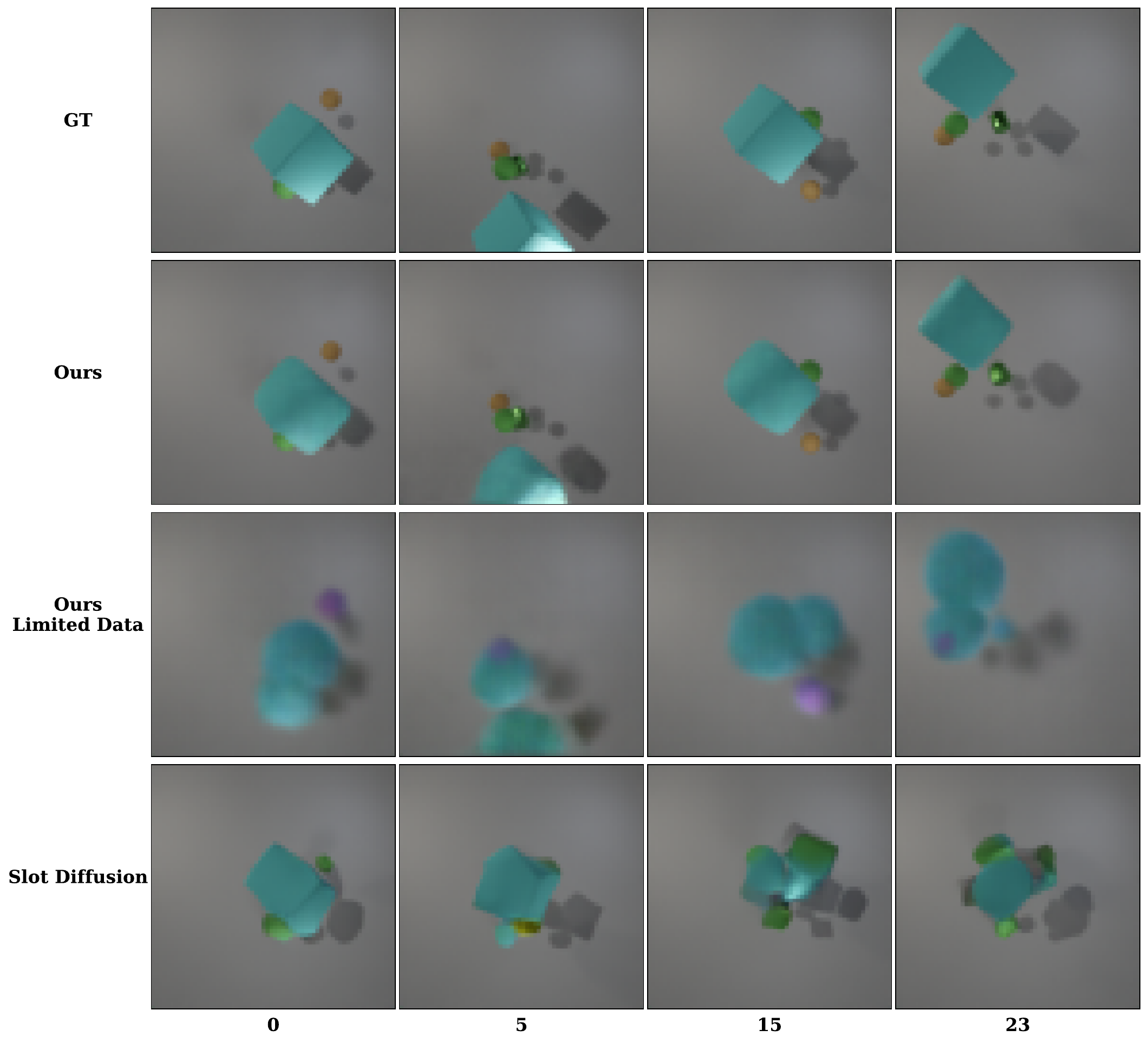}
    \caption{Qualitative performance of different model configurations compared to the ground-truth (GT). Predictions of selected frame iterations are shown from left to right. Our model is able to position objects correctly in future frames, while keeping object shading and overall appearance intact.}
    \label{fig:base_performance}
\end{figure}

For the Orbits dataset we are able to significantly outperform other approaches (cf. \cref{tab:base_performance,fig:base_performance}). Our model is able to follow the correct object trajectories and render the objects accordingly. When comparing the symbolic state $s$ used by $F$ with its auto-encoded version $F_\mathrm{in}(F_\mathrm{out}(s))$ we observe a low \gls*{mae} of $0.005$. This indicates that frame-to-frame object states can be accurately recovered from the latent representation without too much error accumulation.

In order to verify the impact of the integrated procedural knowledge, we train two variants of our model which replace the integrated function $F$ with an identity function, and use only the residual model $R$ in place of our procedural knowledge module $P$ (ablations 3) and 4) in \cref{tab:base_performance}, respectively). In both cases the performance decreases substantially and converges to the same performance as the related work, giving clear evidence that $F$ has a large positive impact on the performance.

Since we expect that the integration of procedural knowledge should decrease the complexity of the learning problem, and as such require less data than a completely uninformed model, we also train our model with a very small dataset consisting of only 300 training samples instead of 10K to test this hypothesis. As can be seen in \cref{tab:base_performance} the performance significantly decreases, but stays well above variants that do not include $F$. When looking at the predictions in \cref{fig:base_performance}, the reason for this is a quality loss in object appearance prediction, however the objects are still rendered at the correct positions. As we will further show in \cref{sec:control}, the latent vectors $z_b$ and $z_c$ contain such appearance features, both of which are purely based on the data-driven aspects of our model. These results indicate that while our data-driven parts suffer equally to regular data-driven models when trained with small amounts of data, the inclusion of a procedural knowledge path $z_a$ can enable the model to still provide meaningful predictions. Through this experiment we thus highlight that our model not only benefits from the integrated knowledge, but also still benefits from additional data, which conforms to our goal of providing domain experts with additional avenues to improve their models without removing options.

We also test our approach in the face of uncertainties in the integrated function. For this we let the model learn the environment constants \texttt{object mass} and \texttt{gravitational constant} present within $F$ in an unsupervised manner. We used a modified learning rate of $1e^{-2}$ instead of the regular $2e^{-4}$ for these parameters to improve and speed up convergence due to the larger magnitudes compared to regular network weights. The performance is comparable to our model without any learned parameters (see ``Learned Parameters in $F$'' in \cref{tab:base_performance}). In fact, the model converges on the ground-truth values of the learned parameters without ever supplying these as a supervision signal, highlighting the potential of our approach for system and parameter identification tasks as well.

Finally, we analyze the impact of individual components of our approach. We train variants of our architecture that do not use the residual path $z_c$ and do not use a residual model $R$ and thus $z_b$ (see ablations 1) and 2) in \cref{tab:base_performance}). Removing either only has a slight negative influence on performance, which aligns with the finding earlier that using $F$ has the largest performance impact in the model. It shows however that predictions can further be improved by using a suitable model $R$, especially when there are environment processes not covered by $F$, such as dynamic lighting conditions.

\subsubsection{Acrobot \& Pendulum Camera}

\begin{table}[t]
    \centering
    \caption{Performance comparison for the Acrobot and the Pendulum Camera dataset. Purely object-centric models such as SlotFormer cannot be applied to the Pendulum Camera dataset as no objects are visible in the scene.}
    \begin{tabular}{p{2.7cm}|l|l|l|l|l|l}
\toprule
& \multicolumn{3}{l|}{\textbf{Acrobot}} & \multicolumn{3}{l}{\textbf{Pendulum Camera}} \\
& \multicolumn{1}{l}{LPIPS$\downarrow$} & \multicolumn{1}{l}{SSIM$\uparrow$} & \multicolumn{1}{l|}{PSNR$\uparrow$} & \multicolumn{1}{l}{LPIPS$\downarrow$} & \multicolumn{1}{l}{SSIM$\uparrow$} & \multicolumn{1}{l}{PSNR$\uparrow$} \\
\midrule
\textbf{Ours} & $\mathbf{3.1}$\textcolor{darkgray}{\scriptsize$\pm0$} &  $\mathbf{97.9}$\textcolor{darkgray}{\scriptsize$\pm0$} &  $\mathbf{37.2}$\textcolor{darkgray}{\scriptsize$\pm2$} & $\mathbf{26.9}$\textcolor{darkgray}{\scriptsize$\pm1$} &  $\mathbf{65.4}$\textcolor{darkgray}{\scriptsize$\pm1$} &  $\mathbf{31.7}$\textcolor{darkgray}{\scriptsize$\pm0$} \\
Slot Diffusion & $18.2$\textcolor{darkgray}{\scriptsize$\pm0$}  & $84.0$\textcolor{darkgray}{\scriptsize$\pm1$} & $26.6$\textcolor{darkgray}{\scriptsize$\pm1$} & N/A& N/A& N/A \\
Takenaka et al. & $4.4$\textcolor{darkgray}{\scriptsize$\pm0$} &  $96.9$\textcolor{darkgray}{\scriptsize$\pm0$} &  $34.9$\textcolor{darkgray}{\scriptsize$\pm0$} & N/A& N/A& N/A \\
Slotformer &    $13.2$\textcolor{darkgray}{\scriptsize$\pm0$} &    $88.0$\textcolor{darkgray}{\scriptsize$\pm0$} &  $28.3$\textcolor{darkgray}{\scriptsize$\pm0$} & N/A & N/A & N/A     \\
PhyDNet    &  $14.2$\textcolor{darkgray}{\scriptsize$\pm2$} &  $90.2$\textcolor{darkgray}{\scriptsize$\pm1$} &  $29.8$\textcolor{darkgray}{\scriptsize$\pm0$} & $50.9$\textcolor{darkgray}{\scriptsize$\pm0$} &  $39.7$\textcolor{darkgray}{\scriptsize$\pm1$} &  $21.7$\textcolor{darkgray}{\scriptsize$\pm1$}     \\
Dona et al. &$31.4$\textcolor{darkgray}{\scriptsize$\pm9$} &  $85.7$\textcolor{darkgray}{\scriptsize$\pm4$} &  $23.7$\textcolor{darkgray}{\scriptsize$\pm8$} & $50.8$\textcolor{darkgray}{\scriptsize$\pm0$} &  $41.6$\textcolor{darkgray}{\scriptsize$\pm2$} &  $21.7$\textcolor{darkgray}{\scriptsize$\pm0$} \\
PredRNN-V2 & $17.5$\textcolor{darkgray}{\scriptsize$\pm2$} &  $88.3$\textcolor{darkgray}{\scriptsize$\pm0$} &  $28.9$\textcolor{darkgray}{\scriptsize$\pm0$} & $50.4$\textcolor{darkgray}{\scriptsize$\pm0$} &  $39.8$\textcolor{darkgray}{\scriptsize$\pm0$} &  $21.8$\textcolor{darkgray}{\scriptsize$\pm0$} \\
\bottomrule
\end{tabular}
    \label{tab:acrobot_performance}
\end{table}

\begin{figure}[t]
\centering
\includegraphics[width=.85\linewidth]{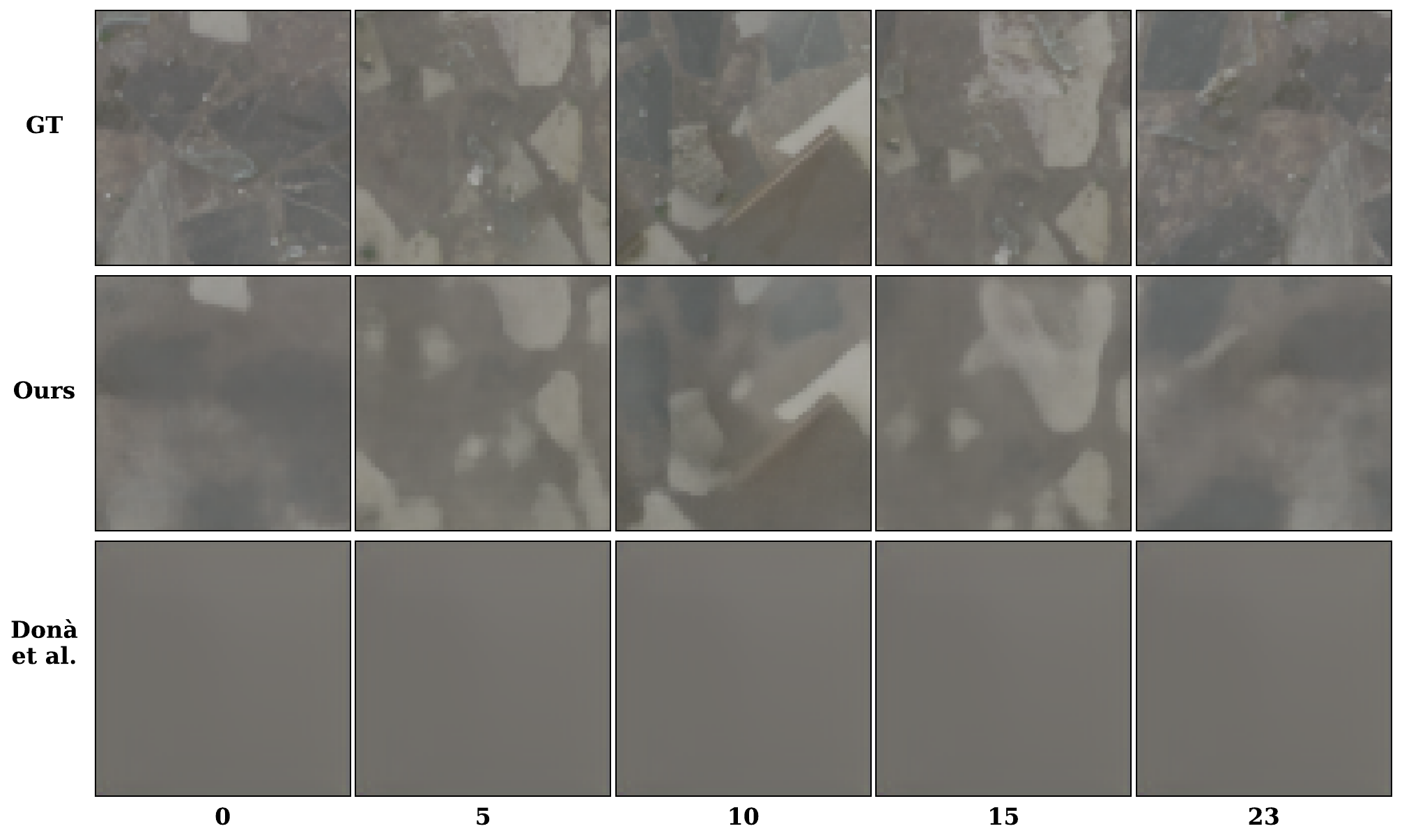}
    \caption{Frame predictions of different time steps (left-to-right) for the Pendulum Camera dataset, with the ground-truth being in the top row (GT).}
    \label{fig:acrobot_performance}
\end{figure}

The Acrobot dataset features a smaller performance gap between our approach and purely data-driven methods such as SlotFormer (cf. \cref{tab:acrobot_performance,fig:acrobot_performance}), indicating that this scenario is easier when relying on data alone. This is probably due to the restricted movements in the scene, whereas in the Orbits setting objects are able to move freely in space, making predictions more difficult. Still, the integration of procedural knowledge was beneficial and lead to improved performance across the board even when compared to Takenaka et al. with a similarly informed model. The challenging nature of modelling indirectly observed dynamics in the Pendulum Camera dataset is reflected in both the quantitative and qualitative performance observed in \cref{tab:acrobot_performance,fig:acrobot_performance}. While our model is able to reconstruct the overall patterns present in each frame, all comparison models were unable to model the pendulum trajectory correctly and instead produced uniform blurry predictions. 

\subsection{Control Interface}\label{sec:control}
The model appears to have used the integrated function $F$ correctly in order to obtain better frame predictions. Since $F$ is based on symbolic inputs and outputs, the question arises whether we can use these to control the predictions in an interpretable manner. To evaluate this, we consider two scenarios: (1) We modify $z_b$ and $z_c$---i.e. the appearance features---and observe whether the dynamics do not change; (2) we directly adjust $z_a$ and observe whether the rendered outputs correlate with our modifications of the symbolic state.

\begin{figure}[t]
\begin{center}
\includegraphics[width=.8\linewidth]{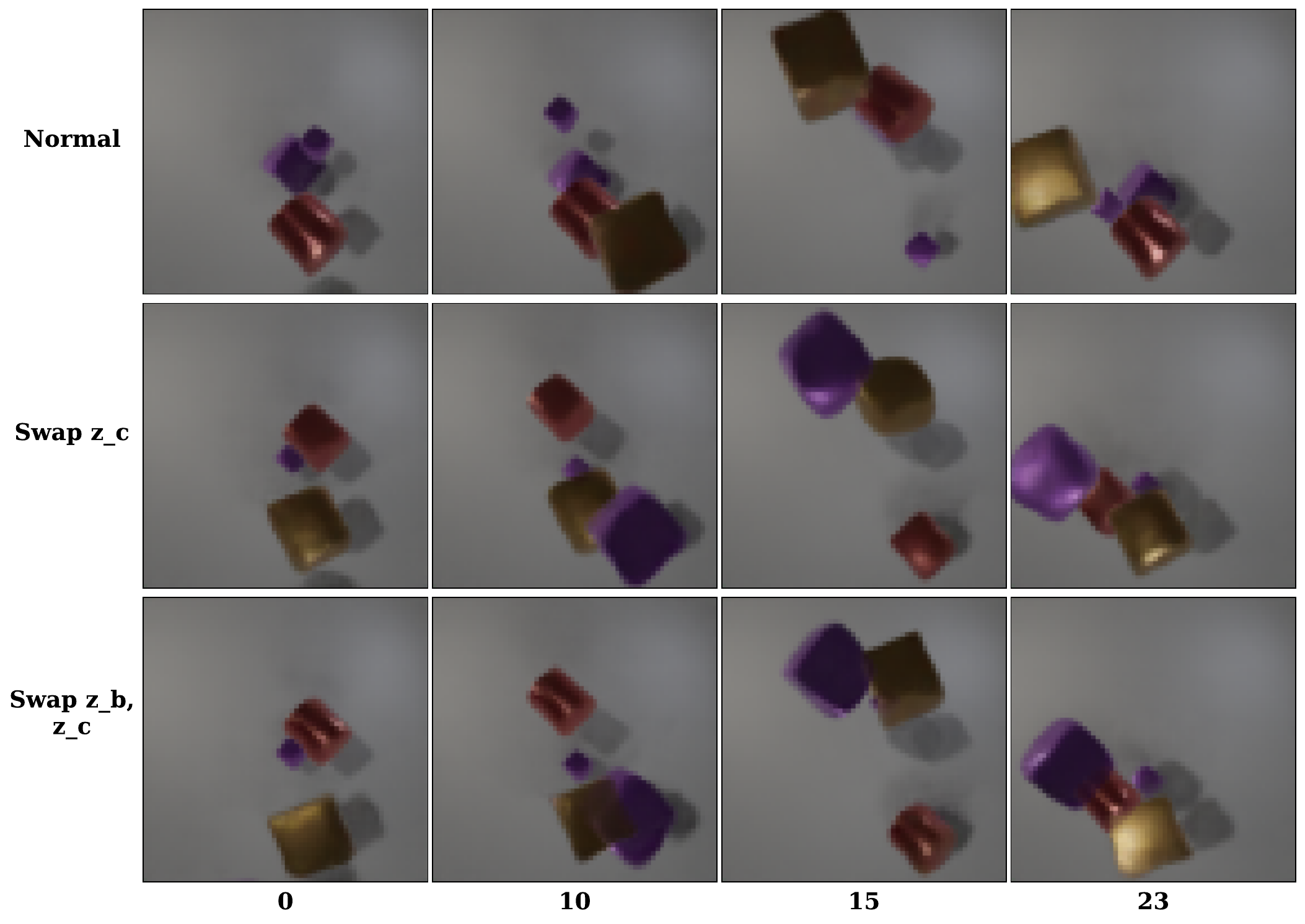}
\caption{Frame predictions of our model for different time steps (left-to-right) in the case of no changes to the latent vector (\textbf{Normal}), swapping $z_c$ between the object latent vectors (\textbf{Swap $\mathbf{z_c}$}) before decoding, and swapping both latent vectors $z_b$ and $z_c$ with the same permutation (\textbf{Swap $\mathbf{z_b, z_c}$}). Object appearances are swapped, but the dynamics stay unchanged.}
\label{fig:gestalt_swap}
\end{center}
\end{figure}

\begin{figure}[t]
\begin{center}
\includegraphics[width=\linewidth]{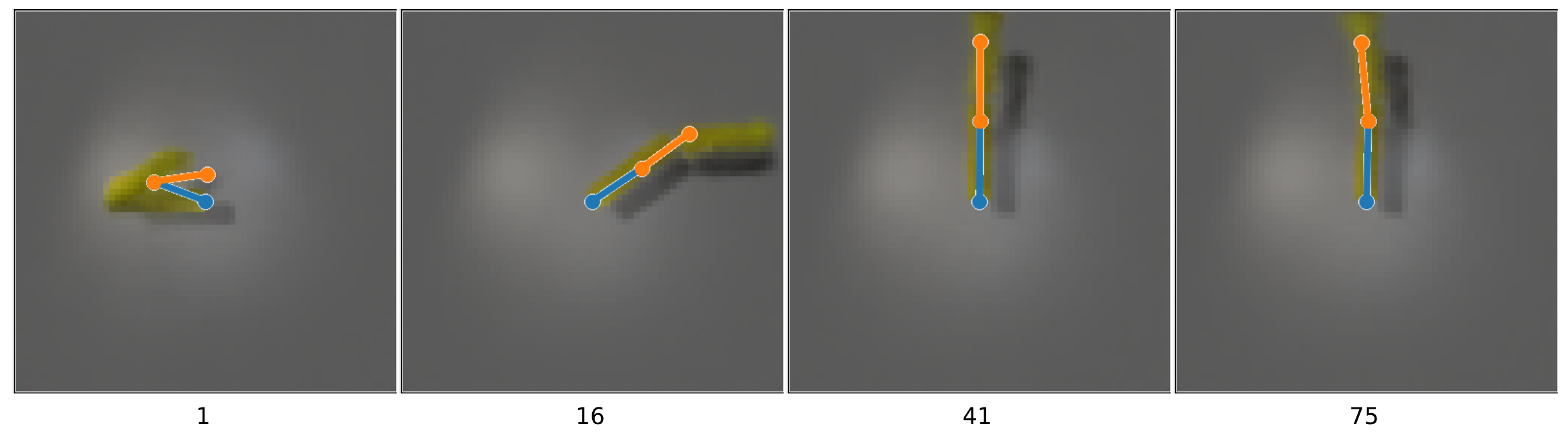}
\caption{Frame predictions of our model for the Acrobot swing-up control task. We overlay a direct plot of the predicted physics states on the rendered frames. Our model renders the states with only slight divergences from the direct plot.}
\label{fig:mpc}
\end{center}
\end{figure}

For (1), we leverage the object-centric representations in the Orbits scenario and simply swap $z_b$ and $z_c$ between different objects. \cref{fig:gestalt_swap} shows the qualitative results. First, as we expected in \cref{sec:arch} we can observe that $z_c$ contains static object properties, such as their size, color, and overall shape. At the same time, when modifying $z_b$ we can see changes in time variant features in the image such as object shading, occlusions, and shape changes due to the cameras perspective projection. Finally, all changes to $z_b$ and $z_c$ did not change the scene dynamics, as the objects simply continue their trajectory, albeit with new appearances.

For (2), we interface our model with \gls*{mpc} and by doing this show that our model not only renders objects at intended locations, but also inherently enables the integration of downstream tasks that operate in symbolic space. We use an off-the-shelf \gls*{mpc} controller that interacts with our model through $F$. We build on top of the Acrobot scenario and define a control task to swing up the double pendulum by allowing a torque to be applied to the joint between the poles. We encode the initial six frames with our model as usual and use the sixth predicted physical state as initialization for the \gls*{mpc} task. We store the predicted torque action sequence for the next 75 frames and apply the corresponding torque for each frame prediction in $F$ before decoding. We show the qualitative results in \cref{fig:mpc}, and more in \cref{apd:mpc}.

We can observe that the model is able to correctly decode predicted object states into a rendered scene. We conclude that the downstream integration of symbolic methods into our method is possible and a large benefit over purely black-box predictors that have no way of interacting in such a manner with existing symbolic tools. We further verified the controllability for the Orbits setting in \cref{apd:test_settings} and show that the model can adapt to new dynamics simply by exchanging $F$.

\section{Limitations and Future Work}\label{sec:limits}
Our work puts a spotlight on the benefit of procedural knowledge represented as programmatic functions and aims to answer some fundamental questions on how it could be utilized in data-driven video predictors. As such, there are still many follow-up questions and directions that can be the topic of future research. 

One assumption that we made was to let the model observe the correct function input of the very first frame in order to give the model a hint of the magnitude and distribution of possible symbolic states. The prediction of initial values---especially in physics---is a known problem with a large body of research behind it. As our focus was on establishing our procedural knowledge module $P$ in the architecture in a domain-independent way, we excluded this aspect as it is part of the encoder module of the underlying model, which should be domain-specific. Still, the integration of a suitable encoder for predicting initial state values is interesting for future work. 

It would also be interesting to combine the given integrated knowledge with synthesis approaches which might dynamically extend or repair the integrated function. Neural program synthesis is an active field of research which might give many insights in this regard. It could also be used to move towards having a dynamically extending library of domain functions available instead of only a single procedural knowledge function, greatly increasing the utility.

\section{Conclusion}\label{sec:conclusion}
We have introduced a novel architectural scheme to join procedural knowledge in the form of programmatic functions with data-driven models. By applying it to video prediction, we show that our approach can enable models to handle tasks for which data-driven models alone would struggle. While our approach also works with very limited data, we highlight that it still benefits from more data, leaving it open for domain experts whether they want to refine their integrated knowledge, or to collect more data, essentially broadening the means by which performance improvements can be made. Furthermore, our grey-box modelling approach increases the transparency of the overall model and allows direct control of the model predictions through the learned interface to the integrated procedural knowledge, enabling easy interfacing with downstream tasks such as \gls*{mpc}.

\bibliographystyle{splncs04}
\bibliography{ControllableVideoGeneration}

\newpage
\appendix

\section{Further Implementation Details}\label{apd:arch}
In the following we describe the core components of our architecture in more detail.

\subsection{Video Frame Encoder}
The used video frame encoder is a standard \gls*{cnn}. The input video frames are encoded in parallel by merging the temporal dimension $T$ with the batch dimension $B$. The \gls*{cnn} consists of four convolutional layers, each with a filter size of 64, kernel size of 5, and a stride of 1. In the non object-centric variant of our architecture, the output features are flattened and transformed by a final fully connected network, consisting of initial layer normalization, a single hidden layer with ReLU activation and a final output linear layer with $C=768$ neurons each. The result is a latent vector of size $B\times T\times C$ that serves as input to $P$.

In the object-centric variant, a position embedding is additionally applied after the \gls*{cnn}, and only the spatial dimensions $H$ and $W$ are flattened before the transformation of the fully connected network, with $C$ reduced to $128$. The result is a latent vector of size $B\times T\times C \times H \times W$. In each burn-in iteration of the object-centric variant, we use the Slot Attention mechanism~\cite{locatelloObjectCentricLearningSlot2020} to obtain updated object latent vectors before applying $P$.

\subsection{Procedural Knowledge Module}
$P$ is responsible for predicting the latent vector of the next frame. It consists of the following submodules:

$P_\mathrm{in}$. Responsible for transforming the latent vector obtained from the image frame encoder into a separable latent vector $z$. It is implemented as a fully connected network with a single hidden layer using the ReLU activation function. All layers have a subsequent ReLU activation function. The number of neurons in all layers corresponds to $C$.

$P_\mathrm{out}$. Responsible for transforming $z$ back into the latent image space. It has the same structure as $P_\mathrm{in}$.

$F_\mathrm{in}$. Responsible for transforming $z_a$ within $z$ into the symbolic space required for $F$. It is a single linear layer without bias neurons. In the object-centric case, its output size directly corresponds to the number of parameters required for $F$ $N_\mathrm{param}$ for a single object. In the non object-centric case when there is no separate object dimension available, it instead corresponds to $N_\mathrm{param} \times N_\mathrm{objects}$, where $N_\mathrm{objects}$ corresponds to the (fixed) number of objects (if present in the dataset).

$F$. Contains the integrated function directly as part of the computational graph. Details about $F$ for the individual data scenarios can be found in \cref{apd:f}.

$F_\mathrm{out}$. Same structure as $F_\mathrm{in}$, with the input and output sizes reversed.

$R$. Responsible for modelling residual dynamics not handled by $F$.  We implement it as a transformer~\cite{vaswaniAttentionAllYou2017} with two layers and four heads. We set the latent size to $C$ and the dimension of its feed-forward network to $512$. It takes into account the most recent 6 frame encodings. Its output corresponds to $z_b$. A temporal position embedding is applied before the transformer.

We first transform the latent image vector into a separable latent vector $z$ by transforming it with $P_\mathrm{in}$. We then split $z$ of size $C$ into the three equally sized components $z_a$, $z_b$, and $z_c$. We continue by obtaining their respective next frame predictions $\hat{z}_a$, $\hat{z}_b$, and $\hat{z}_c$ as follows: $\hat{z}_a$ by $F$, $z_b$ by transforming $z$ with $R$, and $\hat{z}_c$ directly corresponds to $z_c$. All three components are merged back together and transformed into the image latent space with $P_\mathrm{out}$ before decoding.

\subsection{Video Frame Decoder}
We implement the video frame decoder as a Spatial Broadcast Decoder~\cite{wattersSpatialBroadcastDecoder2019b}. We set the resolution for the spatial broadcast to 8, and first apply positional embedding on the expanded latent vector. We then transform the output by four deconvolutional layers, each with filter size 64. We add a final convolutional layer with filter size of 3 to obtain the decoded image. We set the strides to 2 in each layer until we arrive at the desired output resolution of 64, after which we use a stride of 1. In the object-centric variant, we set the output filter size to 4 and use the first channel as weights $w$. We then reduce the object dimension after the decoding as in ~\cite{locatelloObjectCentricLearningSlot2020} by normalizing the object dimension of $w$ via softmax, and using it to calculate a weighted sum with the object dimensions of the RGB output channels.

\subsection{Training Details}
We train all models for at maximum 500k iterations each or until convergence is observed by early stopping. We clip gradients to a maximum norm of $0.05$ and train using the Adam Optimizer~\cite{kingmaAdamMethodStochastic2015} with an initial learning rate of $2e^{-4}$. We set the loss weighting factor $\lambda$ to 1. We set the batch size according to the available GPU memory, which was 32 in our case. We performed the experiments on four NVIDIA TITAN Xp with 12GB of VRAM, taking---on average---one to two days per run.

\section{Details for Comparison Models}\label{apd:relatedwork}
\textbf{Takenaka et al.}~\cite{takenakaGuidingVideoPrediction2023}. We apply the training process and configuration as described in their paper, and instead use RGB reconstruction loss to fit into our training framework. We integrate the same procedural function here as in our model.

\textbf{Slot Diffusion~\cite{wuSlotDiffusionObjectCentricGenerative2023a}}. We use the three-stage training process as described in the paper with all hyperparameters being set as recommended. 

\textbf{SlotFormer~\cite{wuSlotFormerUnsupervisedVisual2023}.} We use their proposed training and architecture configuration for the CLEVRER~\cite{yiCLEVRERCollisionEvents2019} dataset, as its makeup is the most similar to our datasets and follow their proposed training regimen.

\textbf{PhyDNet.} We use their recommended training and architecture configuration without changes.

\textbf{PredRNN-V2.} We use their recommended configuration for the Moving MNIST dataset.

\textbf{Dona et al.}~\cite{donaPDEDrivenSpatiotemporalDisentanglement2021}. We report the performance for their recommended configuration for the Sea Surface Temperature (SST) dataset, as it resulted in the best performance on our datasets.

\section{Further Dataset Details}\label{apd:dataset}
In \cref{tab:dataset} we show further statistics of our introduced datasets.

\begin{table}[h]
    \caption{Detailed statistics of our introduced datasets.}
    \label{tab:dataset}
    \vskip 0.1in
    \centering
    \begin{tabular}{p{.5cm}p{3.5cm}|p{2.5cm}|p{2.5cm}|p{2.5cm}}
         & &\textbf{Orbits}& \textbf{Acrobot} & \textbf{Pendulum Camera}  \\\midrule
         \multicolumn{2}{l|}{Number of} &&&\\
         &training samples & 10K & 2K & 2K\\
         &evaluation samples & 256 & 256 & 256\\
         &burn-in frames & 6 & 6 & 6 \\
         &training rollout frames & 12 & 12 & 12\\
         &validation rollout frames & 24 & 24 & 24\\\midrule
         \multicolumn{2}{l|}{Spatial size} & $64\times64$ & $64\times64$ & $64\times64$\\
         \multicolumn{2}{l|}{Video FPS} & 4 & 10 & 10 \\
         \multicolumn{2}{l|}{Physics FPS} & 40 & 40 & 40 \\
         \multicolumn{2}{l|}{Symbolic State} & Position and velocity of each object & Pole angles and their angular velocities  & As in Acrobot, and also the camera position \\
    \end{tabular}
\end{table}

\newpage
\section{Additional Prediction Visualisations}\label{apd:vis}
\subsection{Orbits}
This section shows additional qualitative results of our model for the Orbits dataset.

\begin{figure}[h]
\centering
    \begin{minipage}{.49\textwidth}
    \includegraphics[width=\columnwidth]{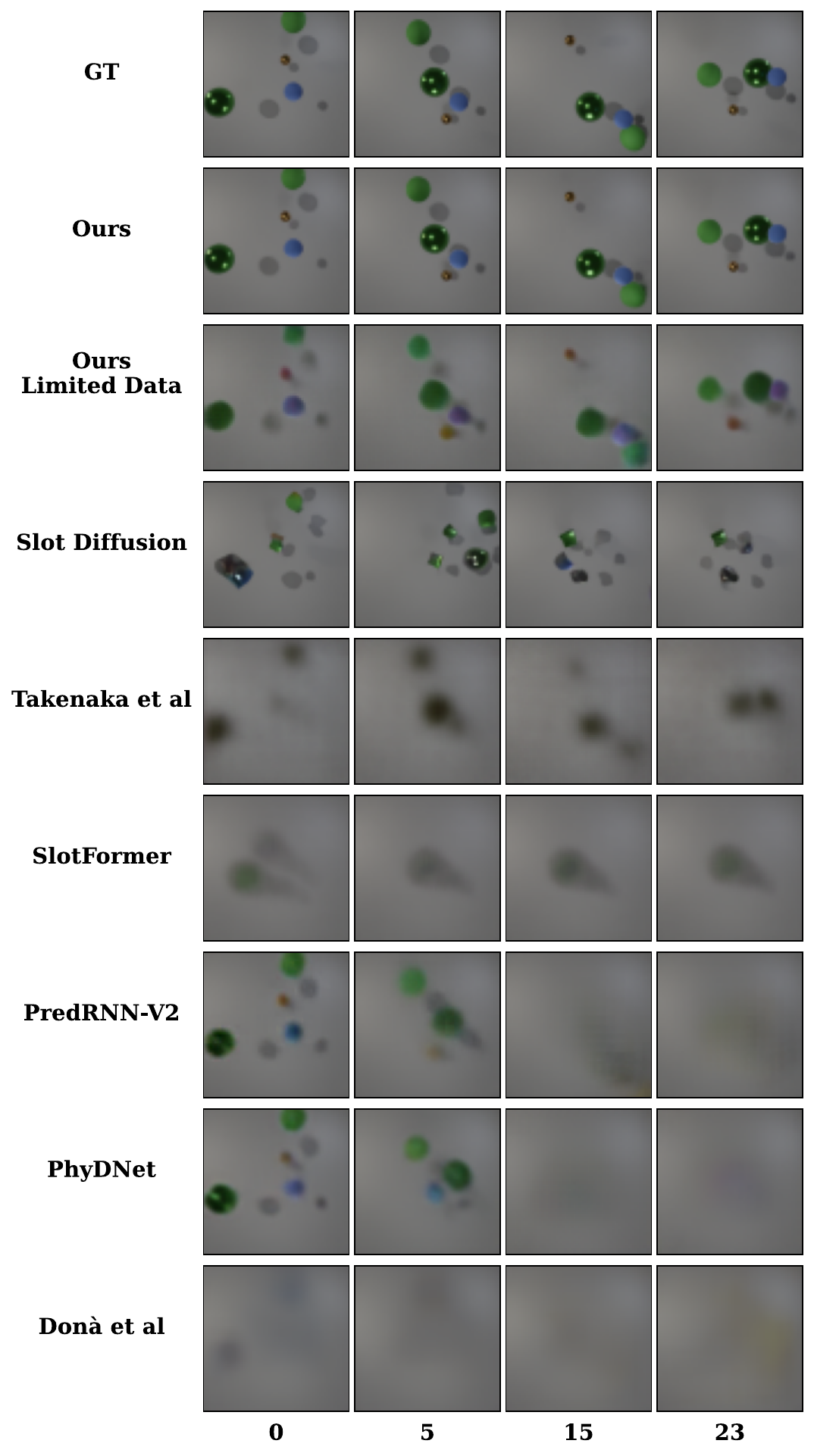}
    \end{minipage}%
    \begin{minipage}{.49\textwidth}
    \includegraphics[width=\columnwidth]{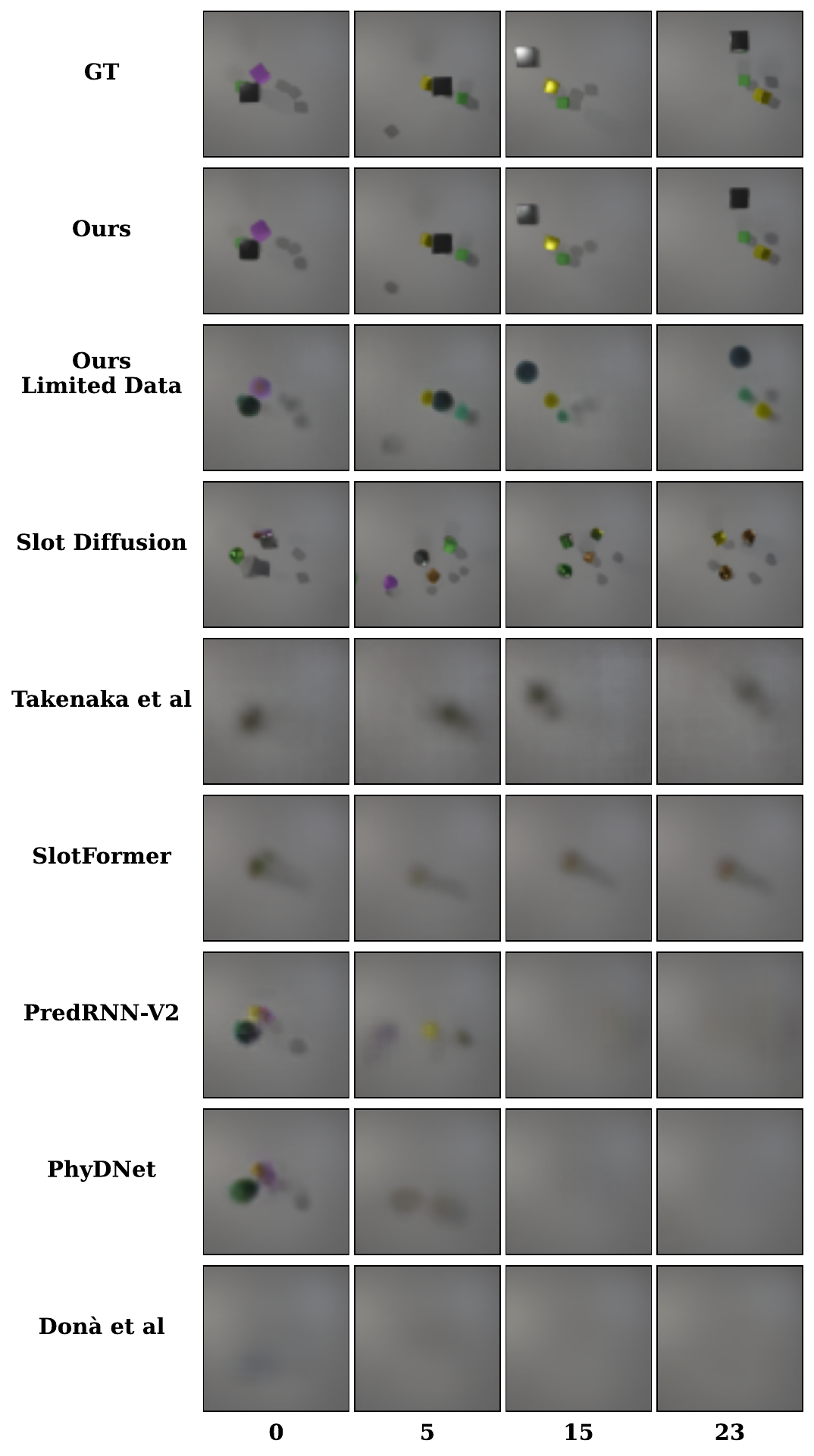}
    \end{minipage}
    \caption{Qualitative results for the Orbits dataset.}
\end{figure}

\newpage

\begin{figure}[h]
\centering
    \begin{minipage}{.5\textwidth}
    \includegraphics[width=\columnwidth]{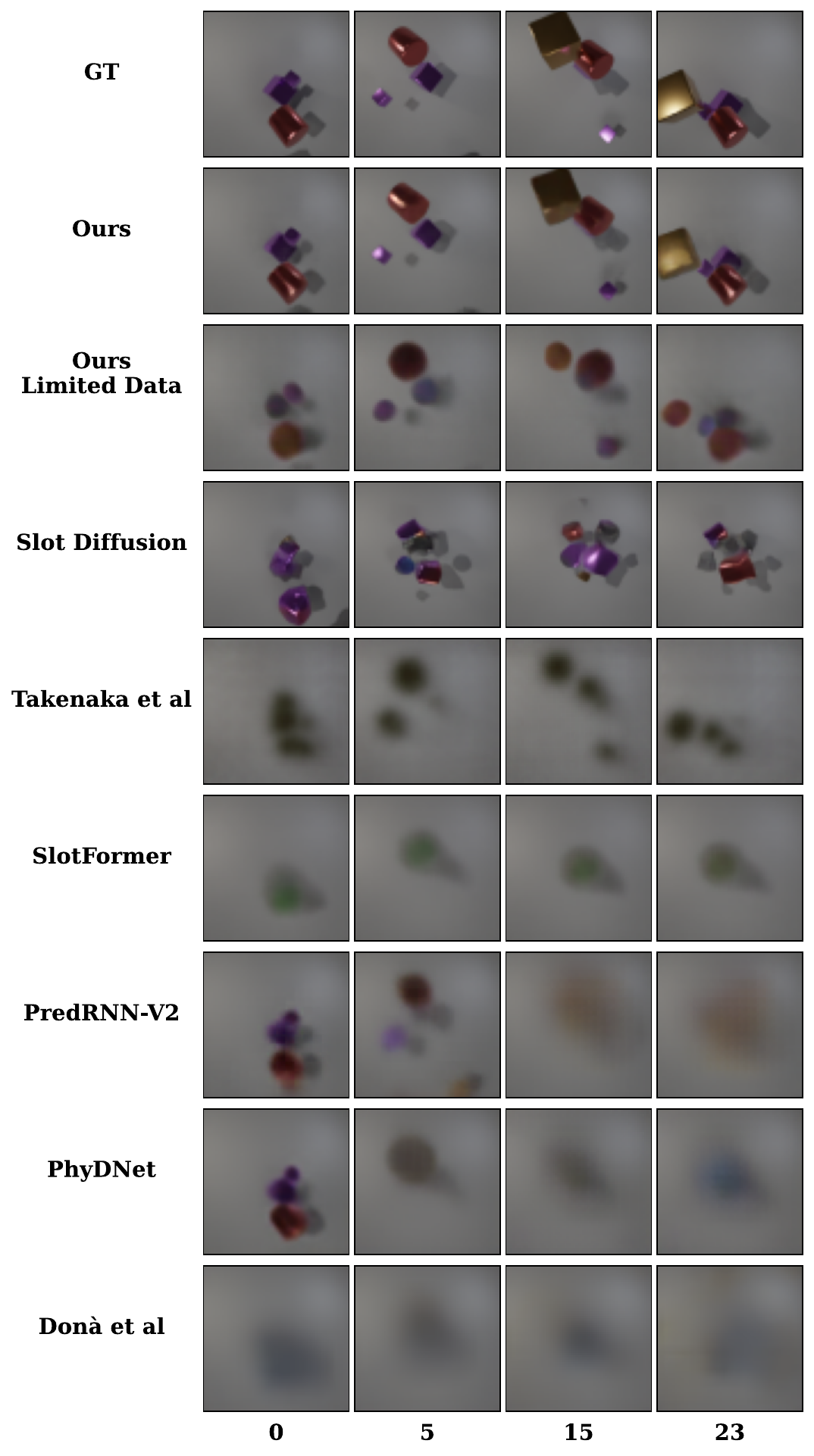}
    \end{minipage}%
    \begin{minipage}{.5\textwidth}
    \includegraphics[width=\columnwidth]{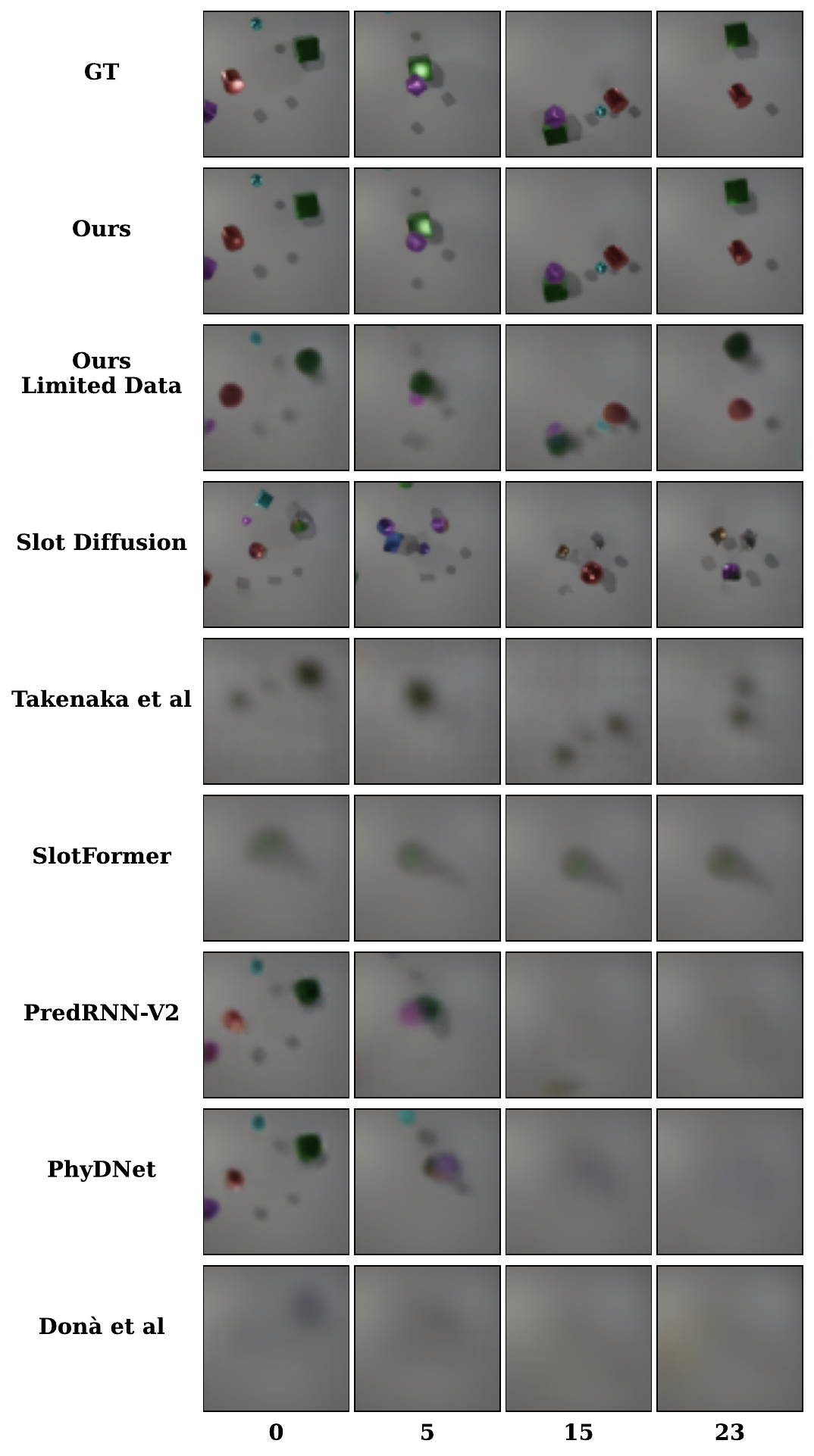}
    \end{minipage}
    \caption{Qualitative results for the Orbits dataset.}
\end{figure}

\newpage

\subsubsection{Latent Vector Swaps}
This section shows additional qualitative results after swapping parts of the latent vector between object representations for the Orbits dataset.

\begin{figure}[h]
\centering
    \begin{minipage}{.40\textwidth}
    \includegraphics[width=\columnwidth]{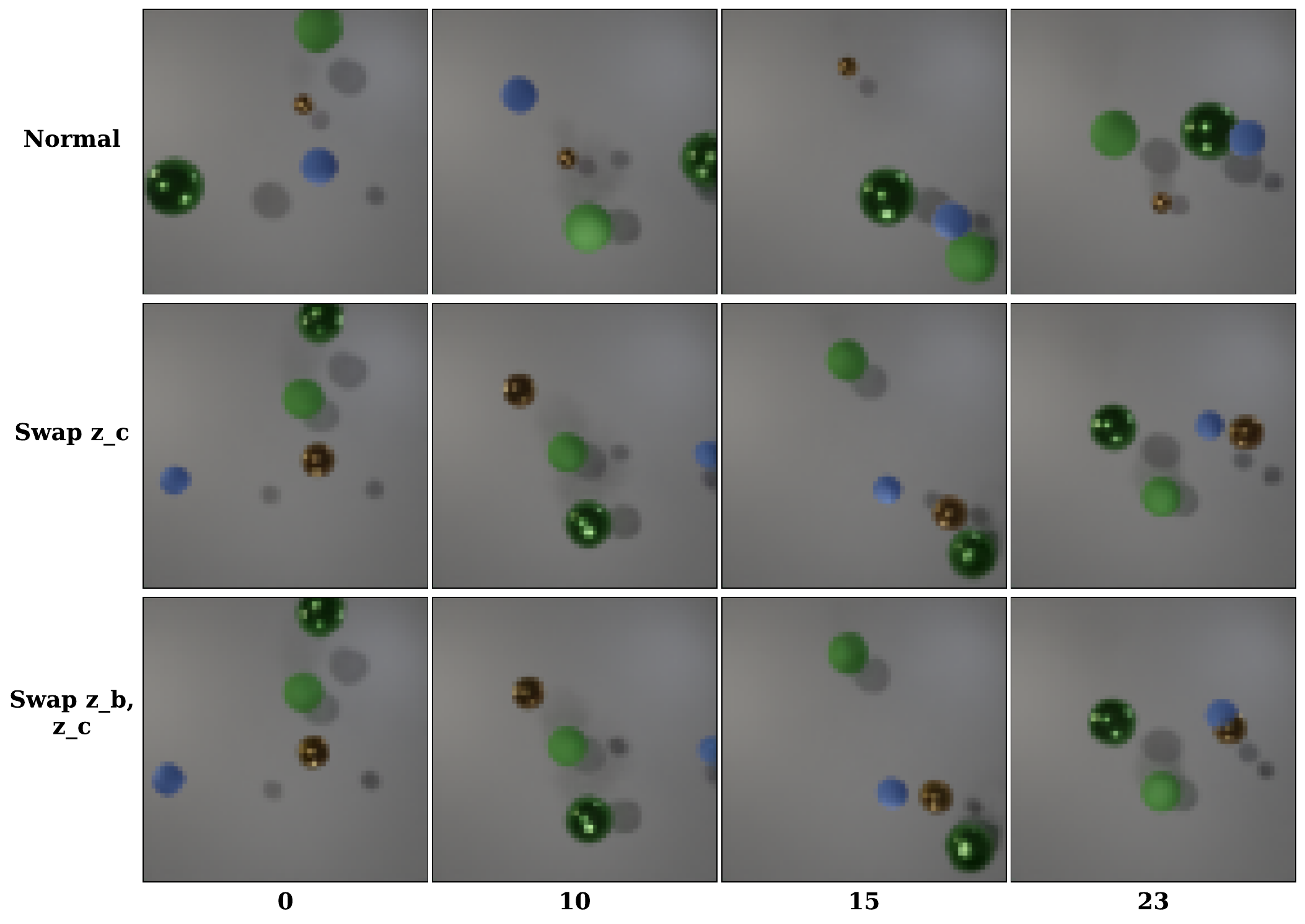}
    \includegraphics[width=\columnwidth]{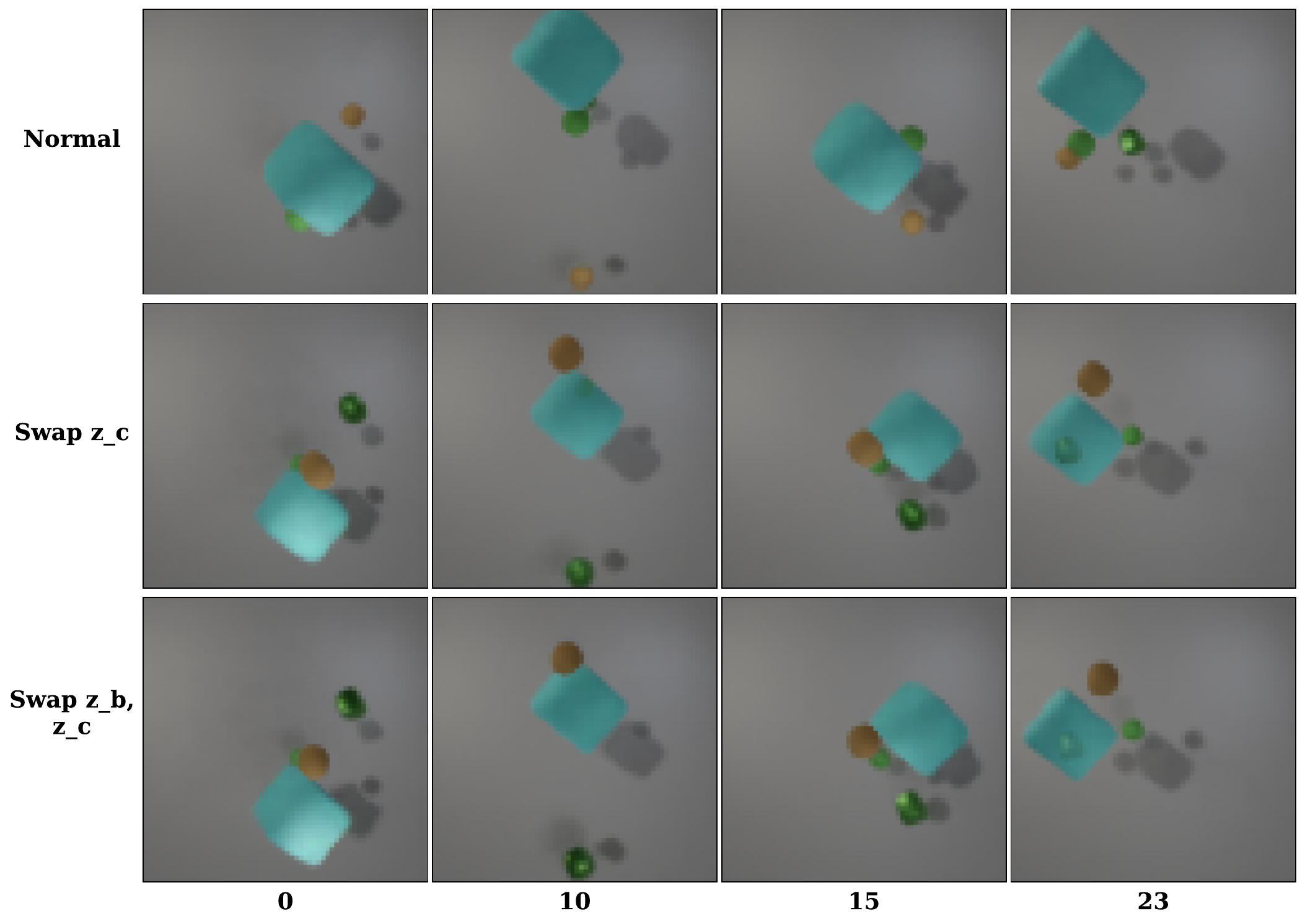}
    \includegraphics[width=\columnwidth]{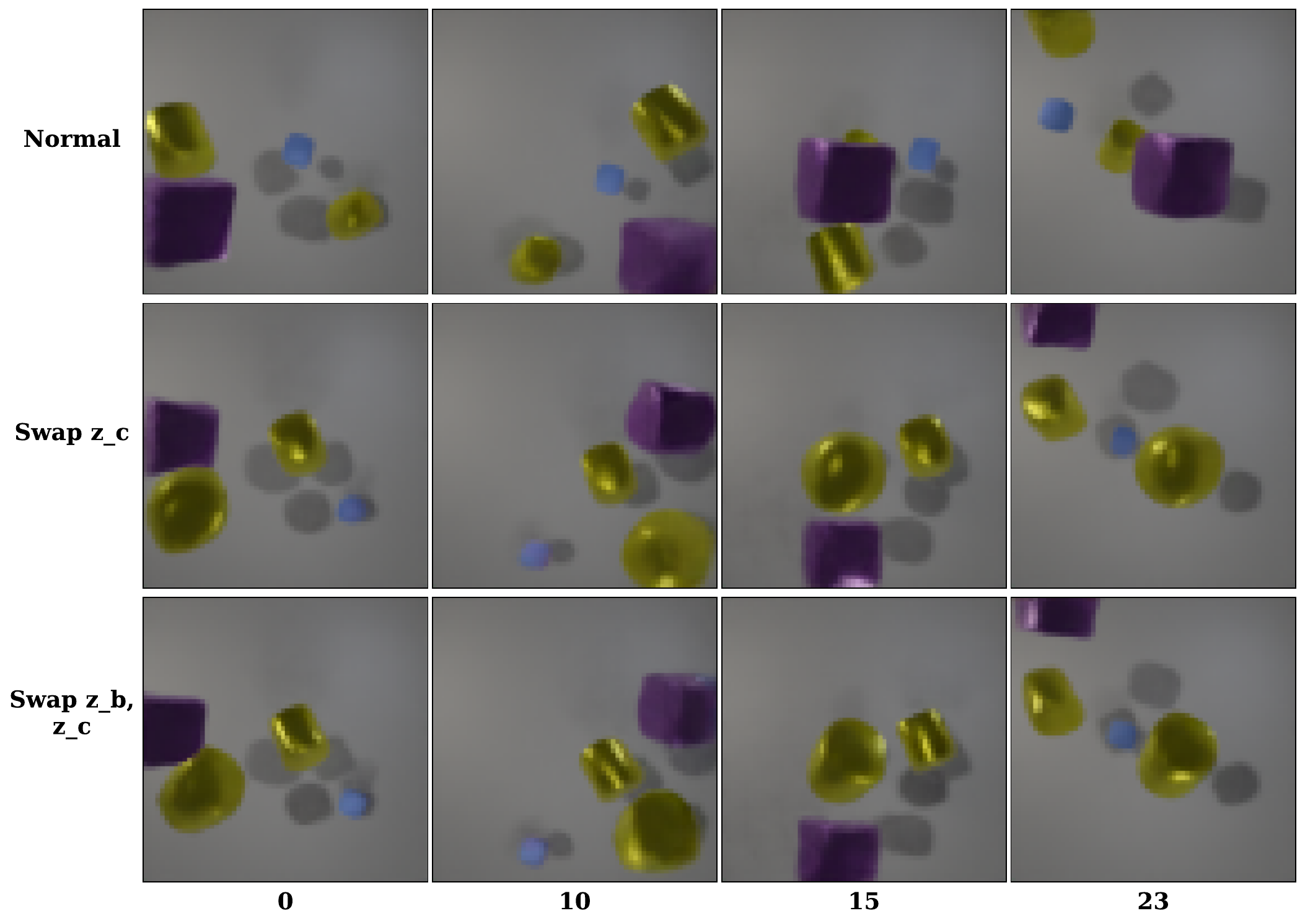}
    \end{minipage}%
    \begin{minipage}{.40\textwidth}
    \includegraphics[width=\columnwidth]{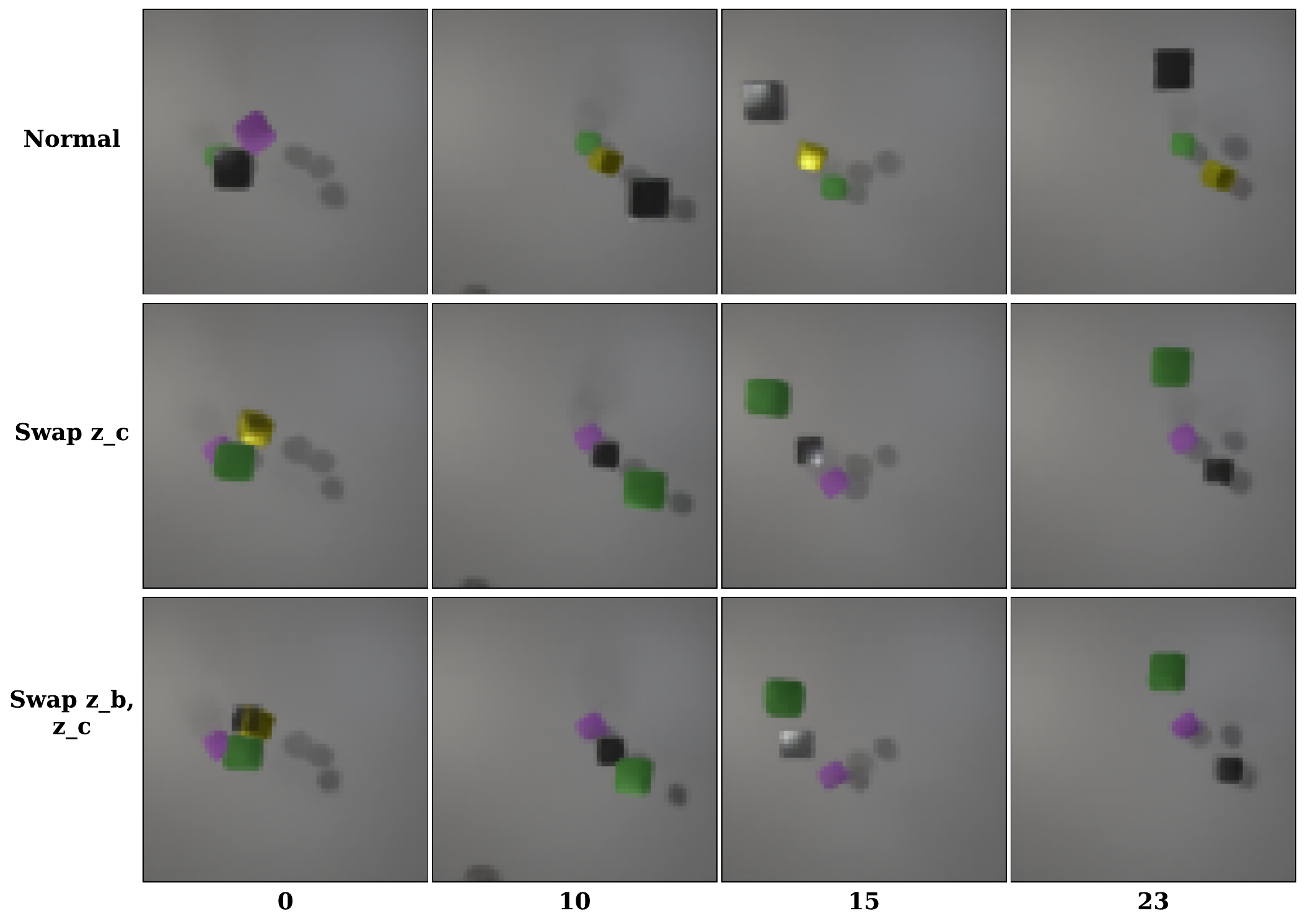}
    \includegraphics[width=\columnwidth]{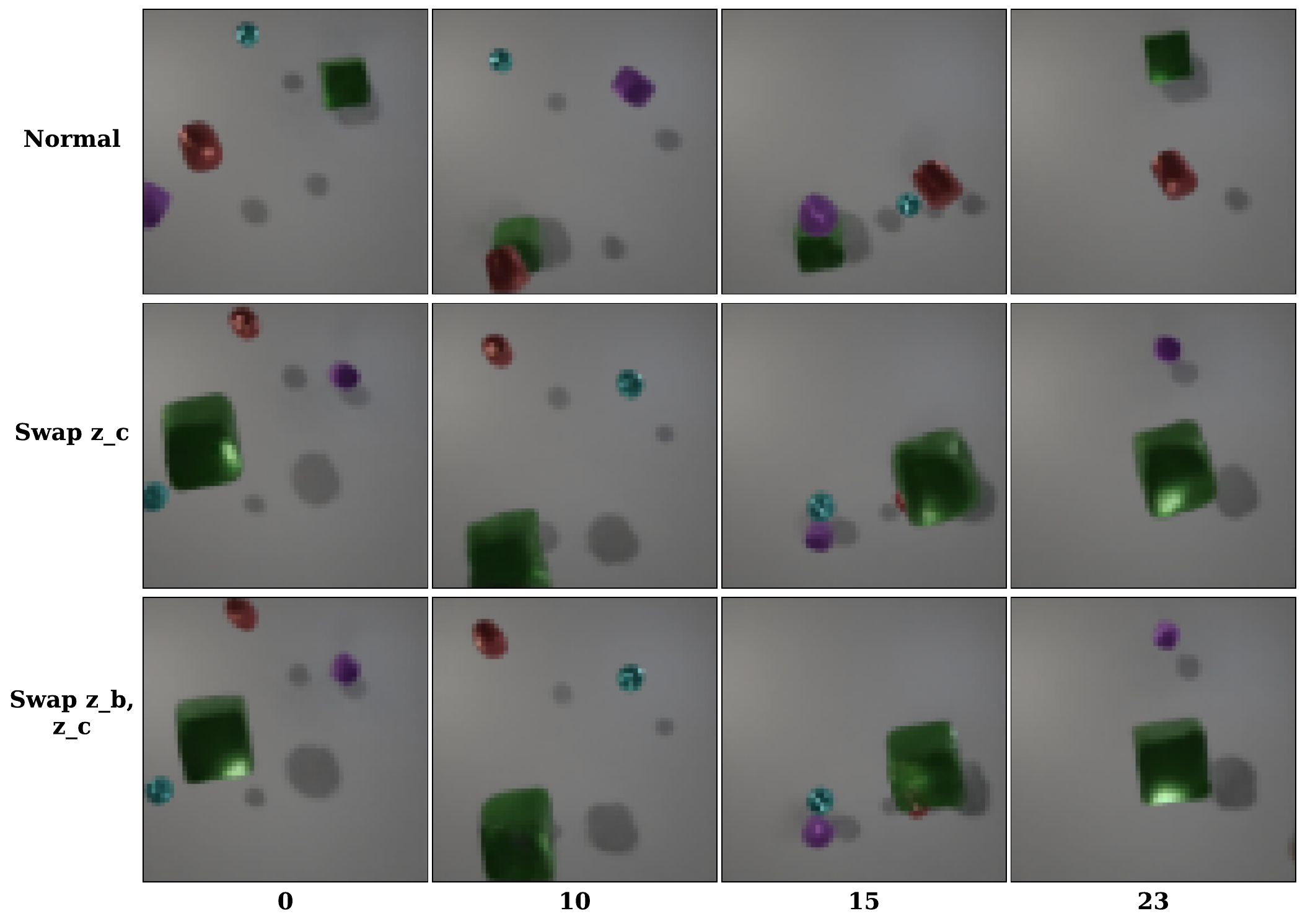}
    \includegraphics[width=\columnwidth]{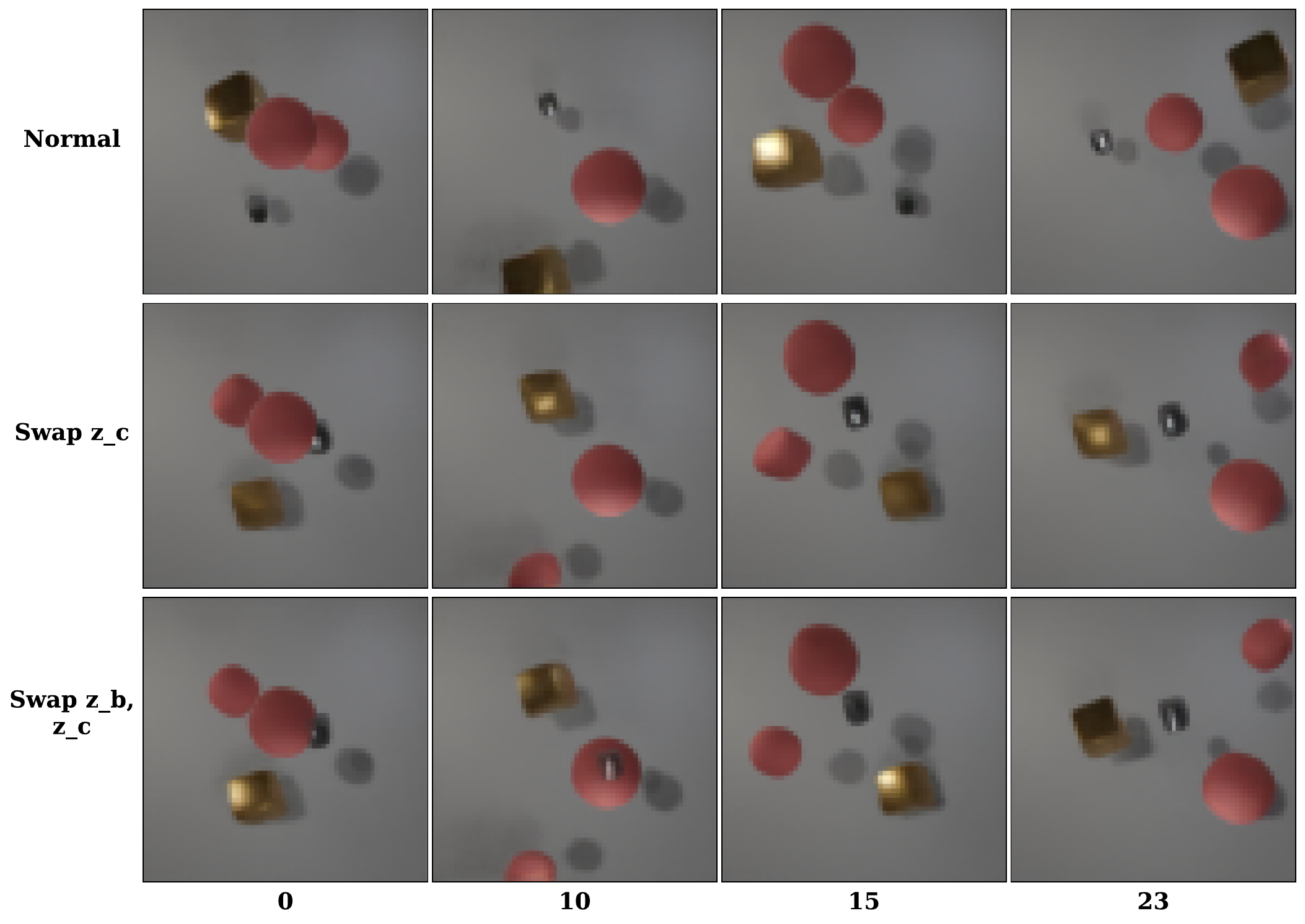}
    \end{minipage}
    \caption{Latent vector swap qualitative result.}
\end{figure}

\newpage

\subsection{Acrobot}
This section shows additional qualitative results of our model for the Acrobot dataset.

\begin{figure}[h]
\centering
    \begin{minipage}{.5\textwidth}
    \includegraphics[width=\columnwidth]{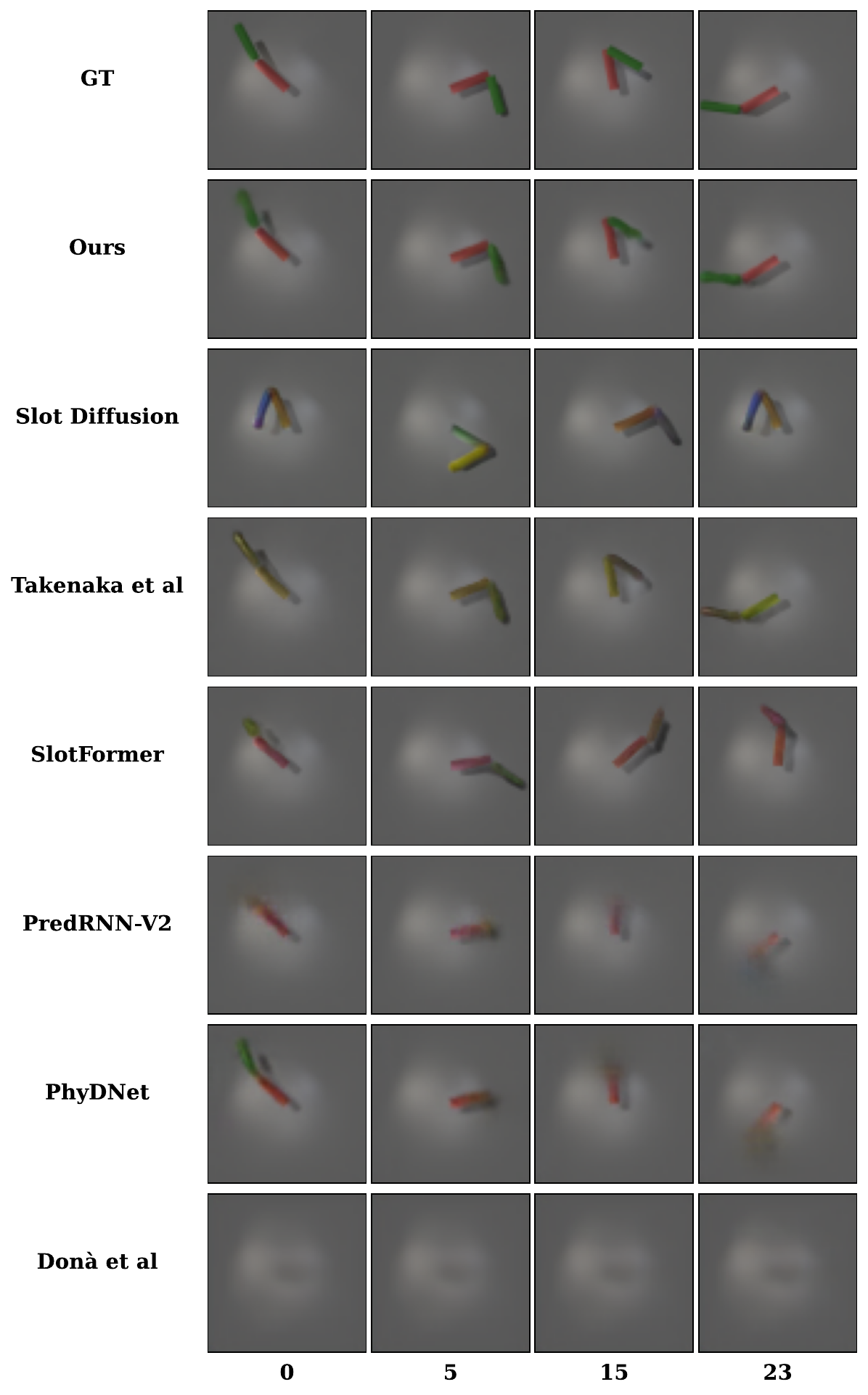}
    \end{minipage}%
    \begin{minipage}{.5\textwidth}
    \includegraphics[width=\columnwidth]{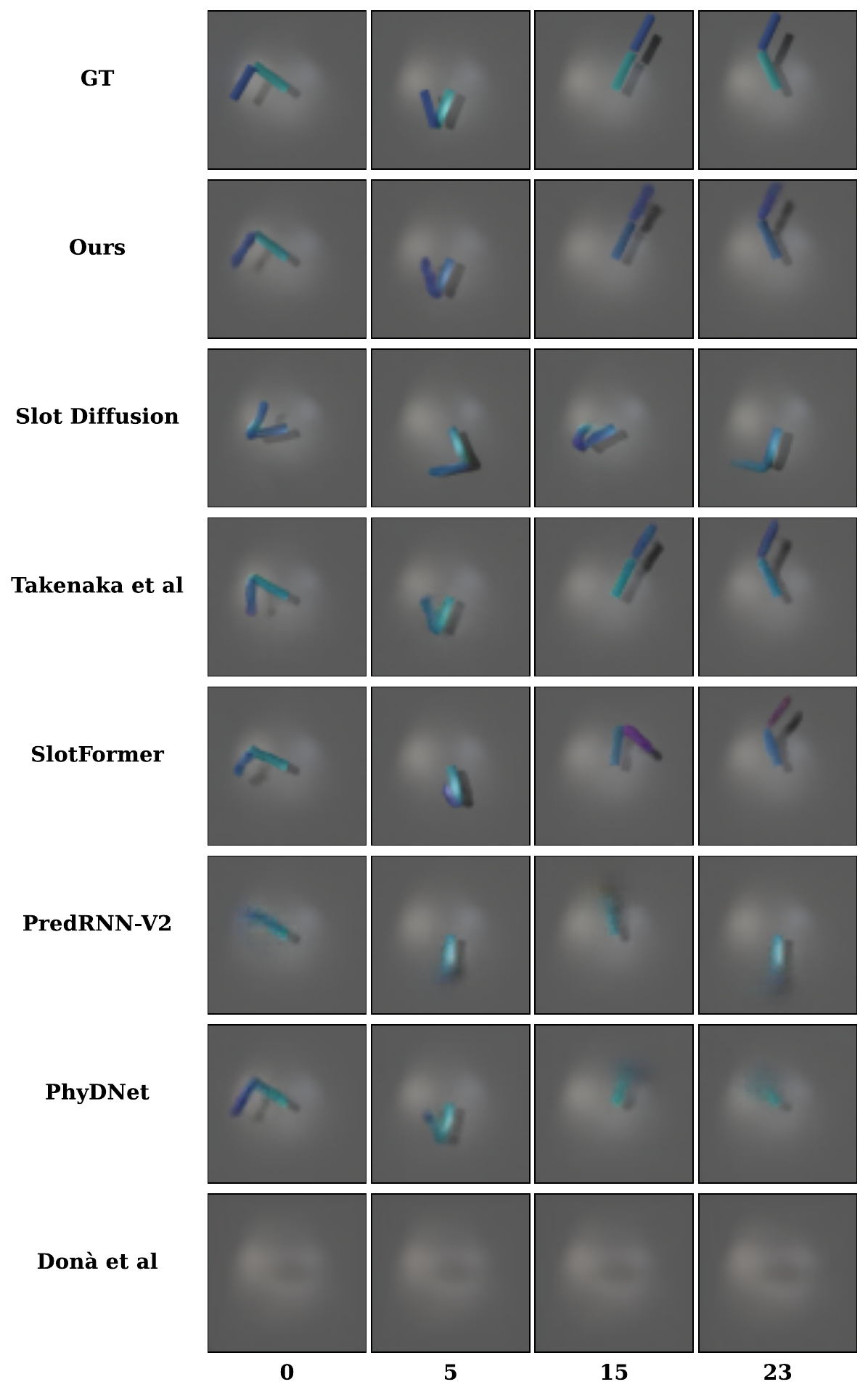}
    \end{minipage}
    \caption{Qualitative results for the Acrobot dataset.}
\end{figure}

\newpage

\begin{figure}[h]
\centering
    \begin{minipage}{.5\textwidth}
    \includegraphics[width=\columnwidth]{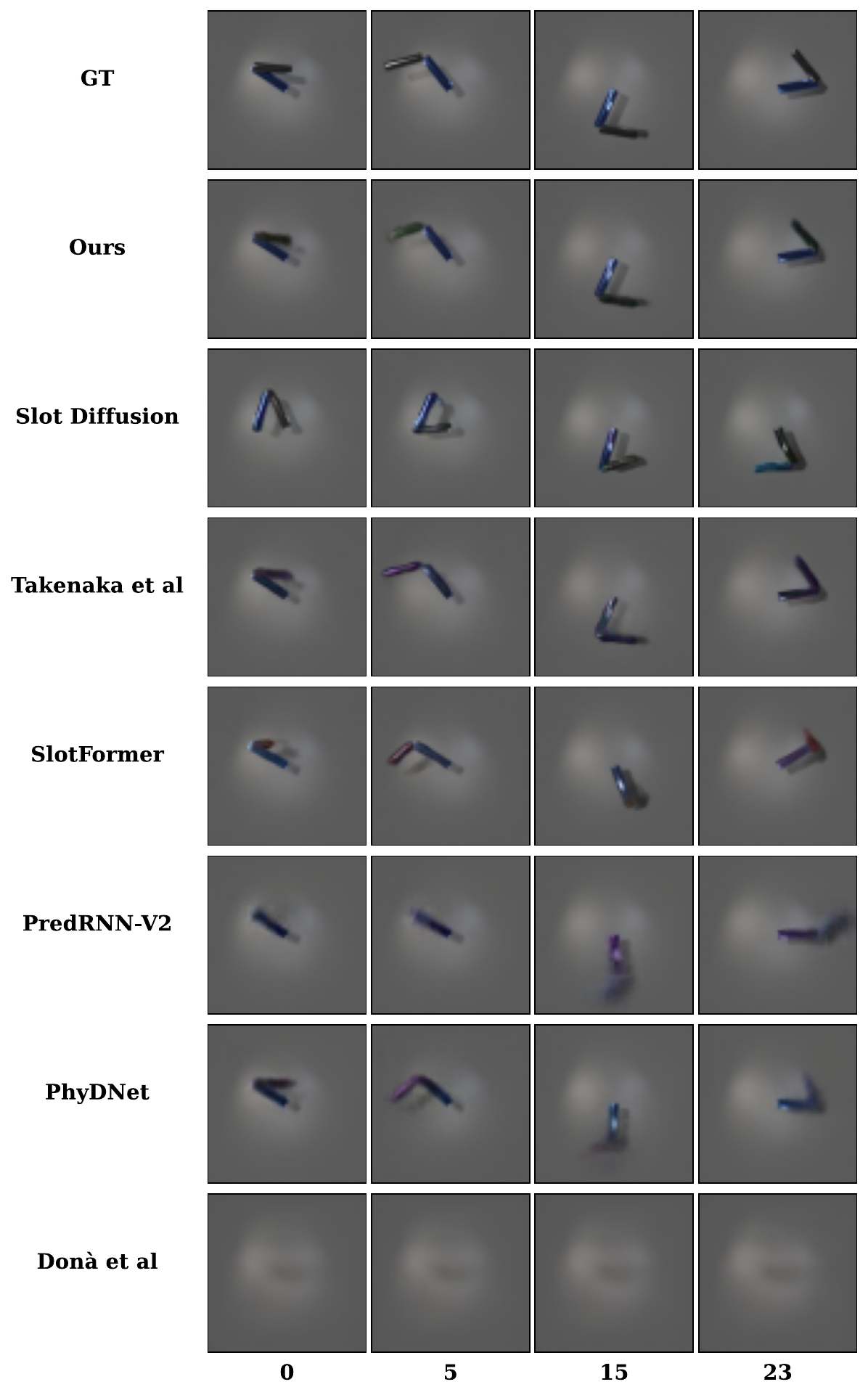}
    \end{minipage}%
    \begin{minipage}{.5\textwidth}
    \includegraphics[width=\columnwidth]{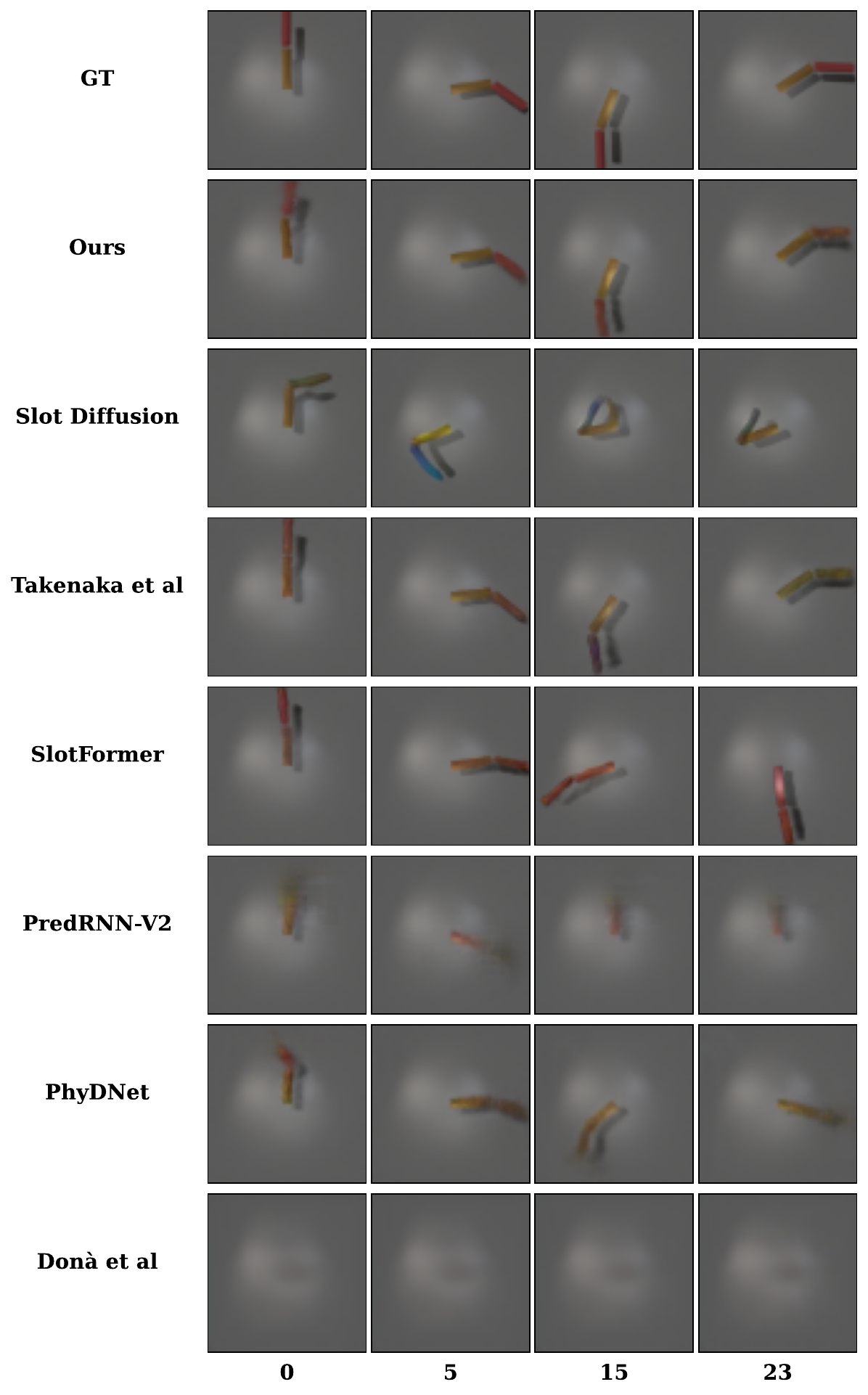}
    \end{minipage}
    \caption{Qualitative results for the Acrobot dataset.}
\end{figure}

\newpage

\subsection{Pendulum Camera}
This section shows additional qualitative results of our model for the Pendulum Camera dataset.

\begin{figure}[h]
\centering
    \begin{minipage}{.46\textwidth}
    \includegraphics[width=\columnwidth]{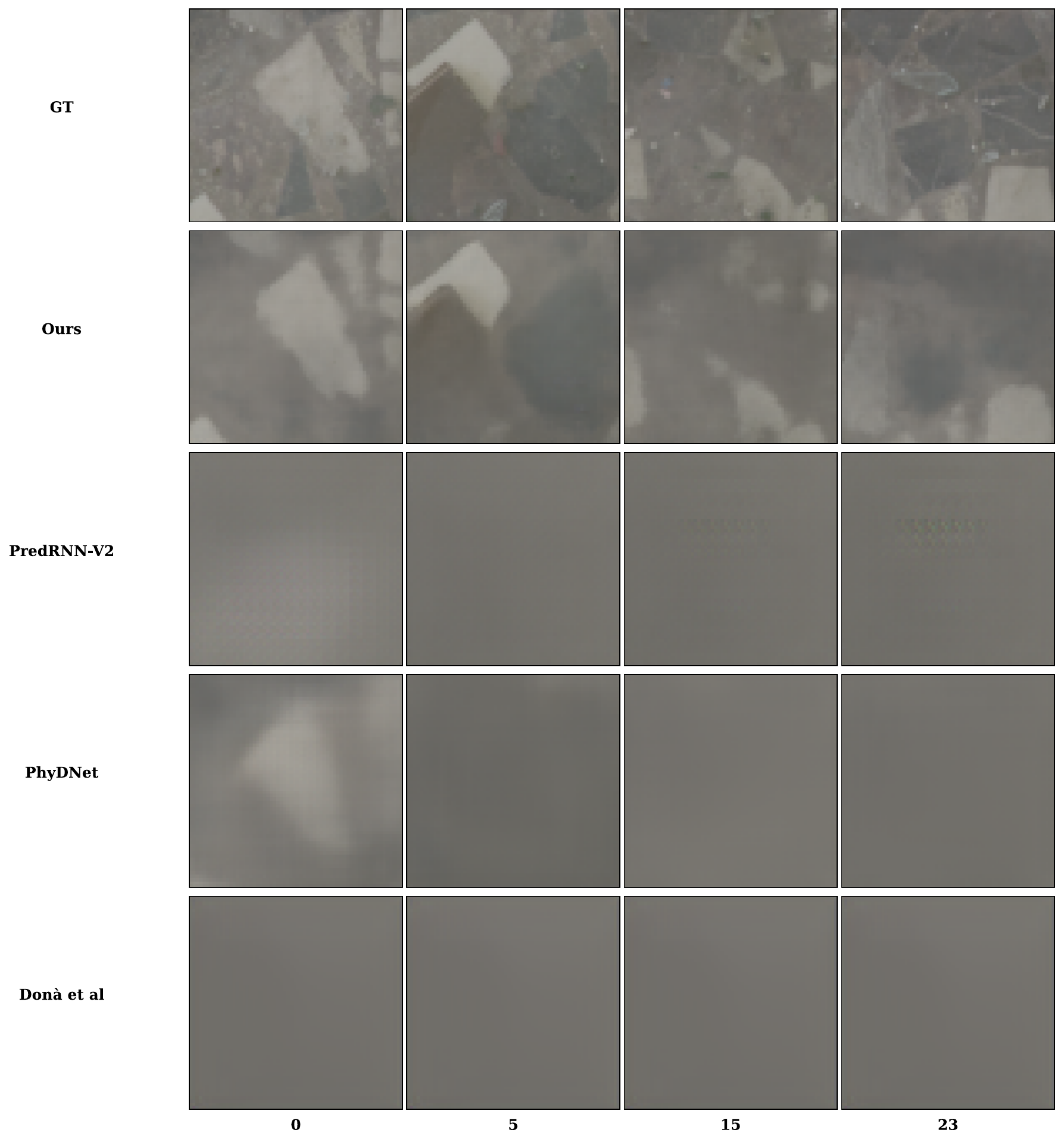}
    \includegraphics[width=\columnwidth]{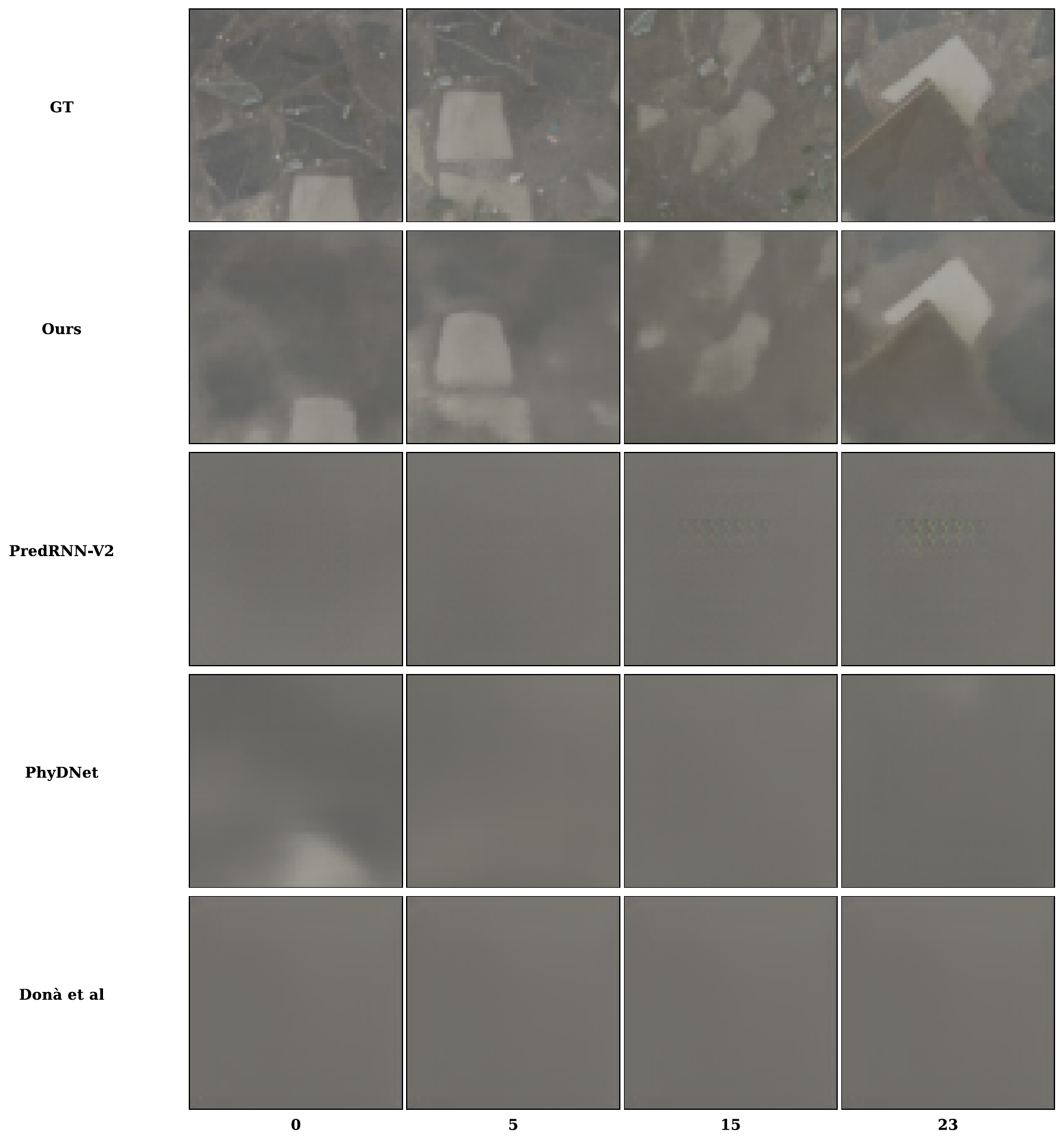}
    \end{minipage}%
    \begin{minipage}{.46\textwidth}
    \includegraphics[width=\columnwidth]{src/2024_01_04_pendulum_vis_2.pdf}
    \includegraphics[width=\columnwidth]{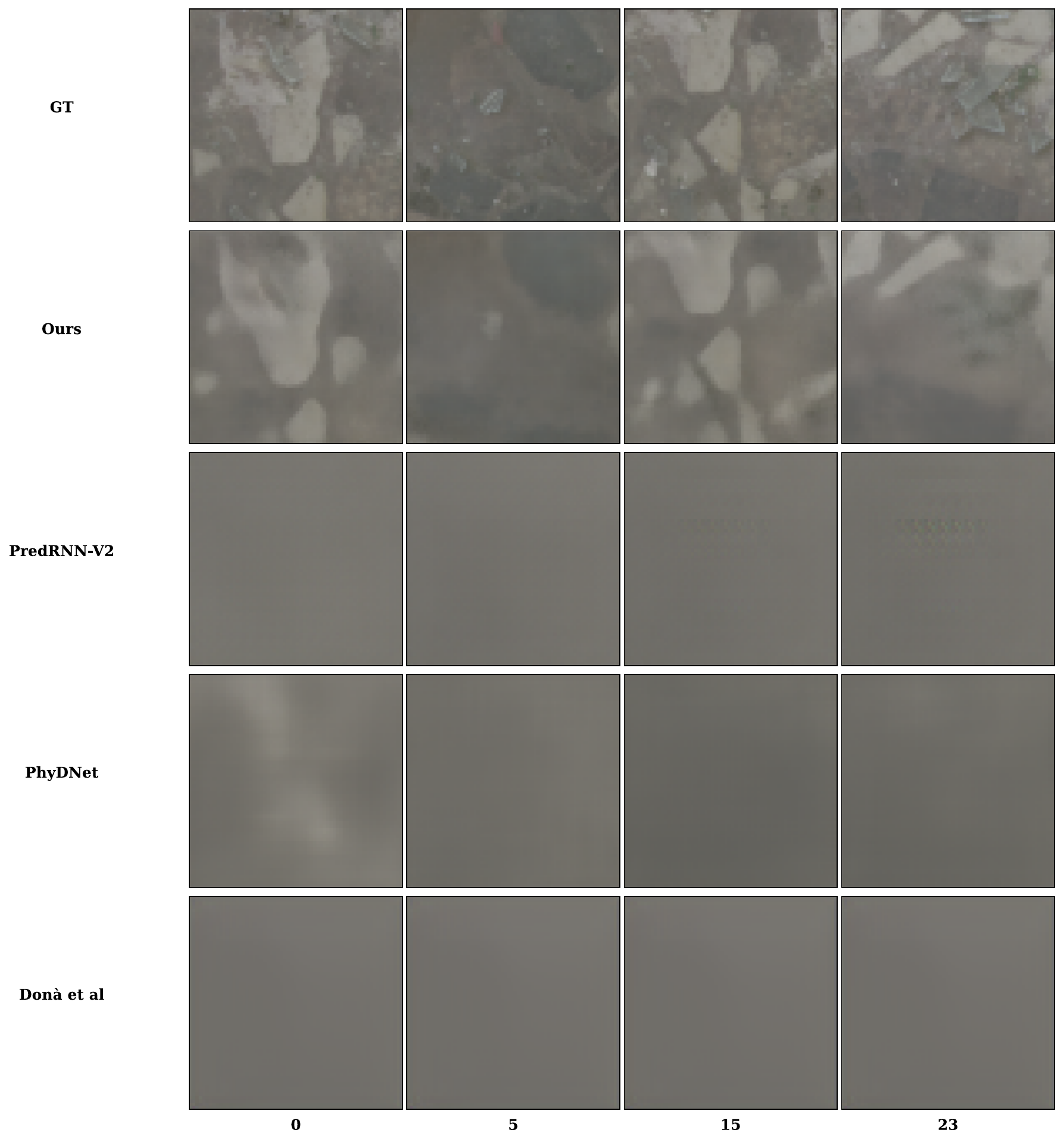}
    \end{minipage}
    \caption{Pendulum Camera dataset qualitative result.}
\end{figure}

\newpage

\section{Orbits Control Validation Dataset Details}\label{apd:test_settings}
In the Orbits setting the object positions are part of the symbolic state, which are an integral factor of correctly rendering the output frame. However, it is not trivial to measure how well our model is able to decode ``hand-controlled'' 3D object positions into a 2D frame in a generalizable manner. Therefore we chose to setup an empirical evaluation framework by assembling variations of the Orbits dataset, ranging from different simulation parameters, over completely novel dynamics, up to non-physics settings such as trajectory following. For each validation set, we replace $F$ of a model trained on the default Orbits dataset with the respective version that handles these new dynamics, and then validate the model without any retraining. 

\begin{table}[h]
    \caption{LPIPS$\downarrow$ Performance on the default Orbits dataset and the validation settings \textbf{A-H}. \textbf{A}: Increased frame rate; \textbf{B}: Increased gravitational constant; \textbf{C}: Tripled force; \textbf{D}: Repulsion instead of attraction; \textbf{E}: No forces; \textbf{F}: No forces and zero velocities; \textbf{G}: Objects follow set trajectories; \textbf{H}: Objects appear at random locations in each frame.}
    \label{tab:test_setting_performance}
    \begin{center}
    \begin{tabular}{p{1.59cm}|l|l|l|l|l|l|l|l}
\toprule
              \textbf{Default}& \textbf{A} & \textbf{B}& \textbf{C} & \textbf{D} & \textbf{E} & \textbf{F} & \textbf{G}& \textbf{H} \\
\midrule
                                    5.6 &                    5.9 &                    5.8 &                    5.8 &                    5.1 &                    2.4 &                    5.4 &                    4.5 &                    4.7 \\
\bottomrule
\end{tabular}
\end{center}
\end{table}

As can be seen in \cref{tab:test_setting_performance}, the performance across all validation settings is comparable to the default dataset and thus, shows that the outputs of $F$ work as a reliable control interface at test time. We note that the much lower \gls*{lpips} for test setting E is due to the objects quickly leaving the scene, resulting in mostly background scenes.

\newpage

\subsection{Dataset Details}
This section shows more details about the Orbits dataset variants that are used to empirically estimate the control performance. \cref{tab:test_settings} describes each setting in more detail, and qualitative results are shown in \cref{sec:test_setting_qualitative}.

\begin{table}[h]
    \caption{Detailed description of the Orbits dataset variants that are used to verify generative control over the integrated function parameters.}
    \label{tab:test_settings}
    \begin{center}
    \begin{tabular}{p{1cm} p{1cm} p{8cm}}
    \toprule
         \textbf{Setting} & & \textbf{Description} \\\midrule
         \multicolumn{2}{l}{Original Dynamics} & \\
         & \textbf{A} & Increased frame rate from 4 to 10 frames per second \\\cmidrule(l){3-3}
         & \textbf{B} & Increased gravitational constant for the physical simulation from 7.0 to 20.0 \\\cmidrule(l){3-3}
         & \textbf{C} & Tripled the force applied to objects\\\midrule
         \multicolumn{2}{l}{Novel Dynamics} & \\
         & \textbf{D} & Objects are repulsed instead of attracted to each other by inverting the force. However, they are still attracted to the camera focal point in order to stay in view longer \\\cmidrule(l){3-3}
         & \textbf{E} & No force is applied to the objects, however initial velocities are kept \\\cmidrule(l){3-3}
         & \textbf{F} & No force is applied to the objects. In addition, all objects have zero velocity, i.e. no object moves \\\midrule
         \multicolumn{2}{l}{Non-Physical Dynamics} & \\\cmidrule(l){3-3}
         & \textbf{G} & Objects move along a predefined 6 point trajectory \\\cmidrule(l){3-3}
         & \textbf{H} & Objects are at a randomly sampled location in each frame \\\bottomrule
    \end{tabular}
    \end{center}
\end{table}

\newpage

\subsection{Qualitative Results}\label{sec:test_setting_qualitative}
This section shows qualitative results for the Orbits validation dataset variants.

\begin{figure}[H]
\centering
    \begin{minipage}{.46\textwidth}
    \includegraphics[width=\columnwidth]{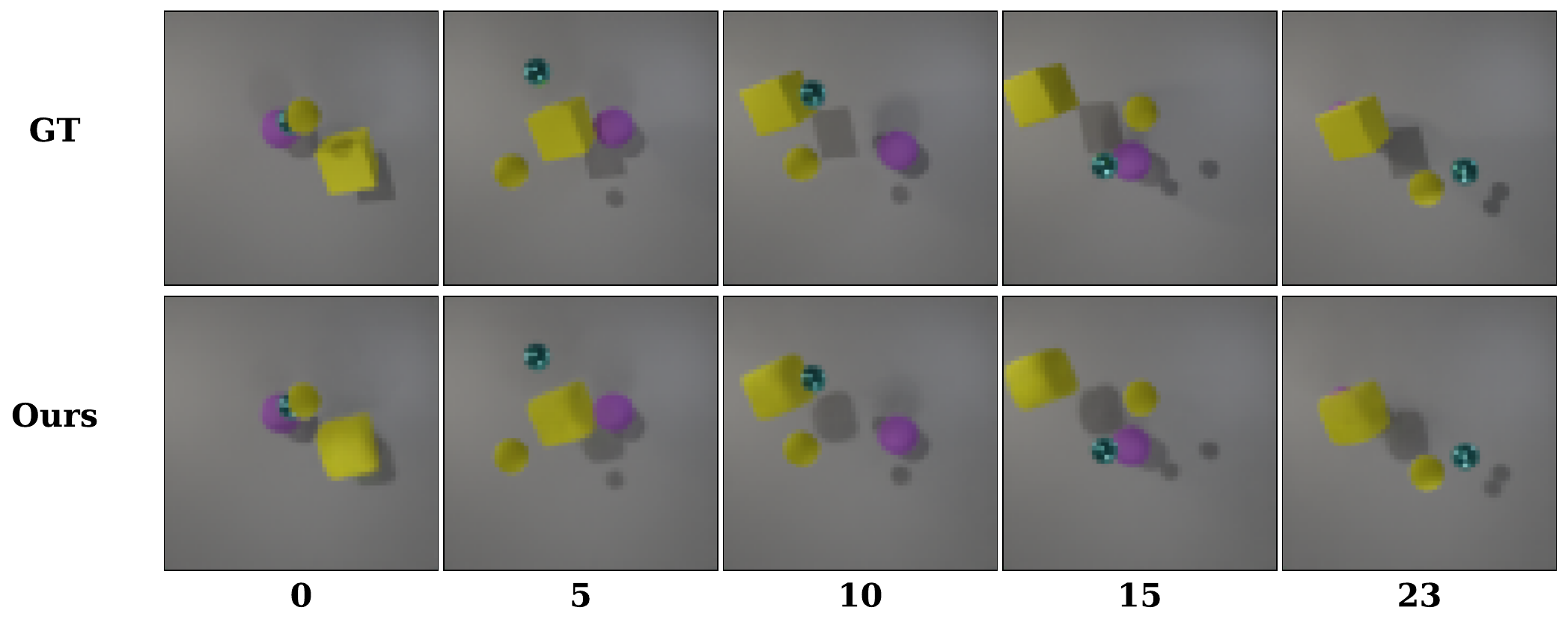}
    \end{minipage}%
    \begin{minipage}{.46\textwidth}
    \includegraphics[width=\columnwidth]{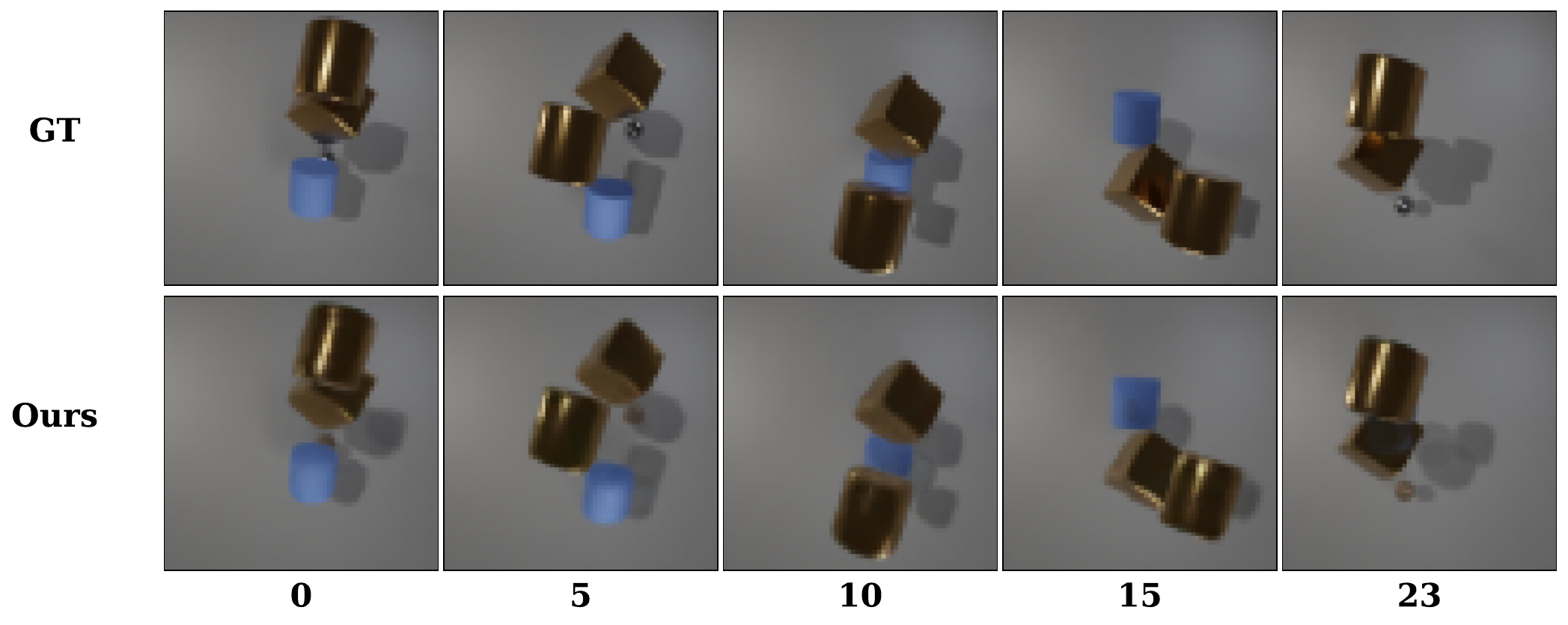}
    \end{minipage}
    \caption{Qualitative performance of our model for the dataset variant A: Increased frame rate}
\end{figure}

\begin{figure}[H]
\centering
    \begin{minipage}{.46\textwidth}
    \includegraphics[width=\columnwidth]{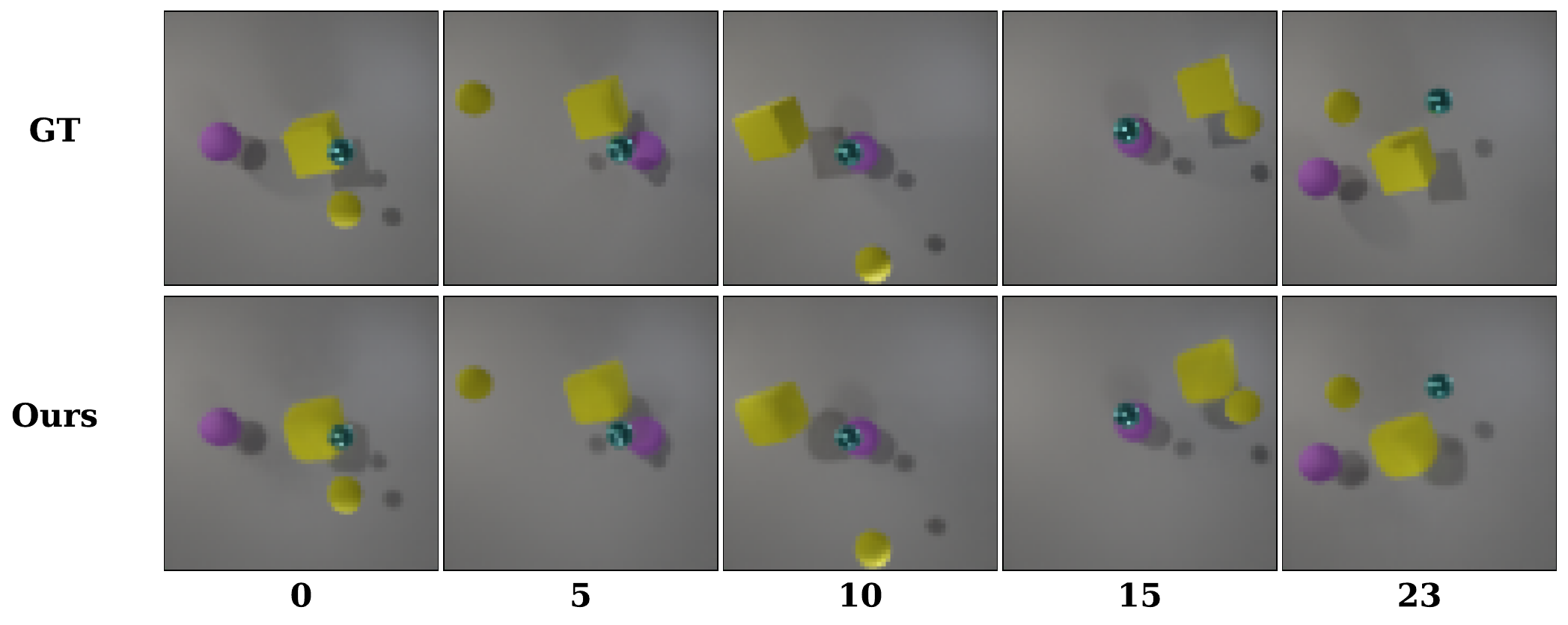}
    \end{minipage}%
    \begin{minipage}{.46\textwidth}
    \includegraphics[width=\columnwidth]{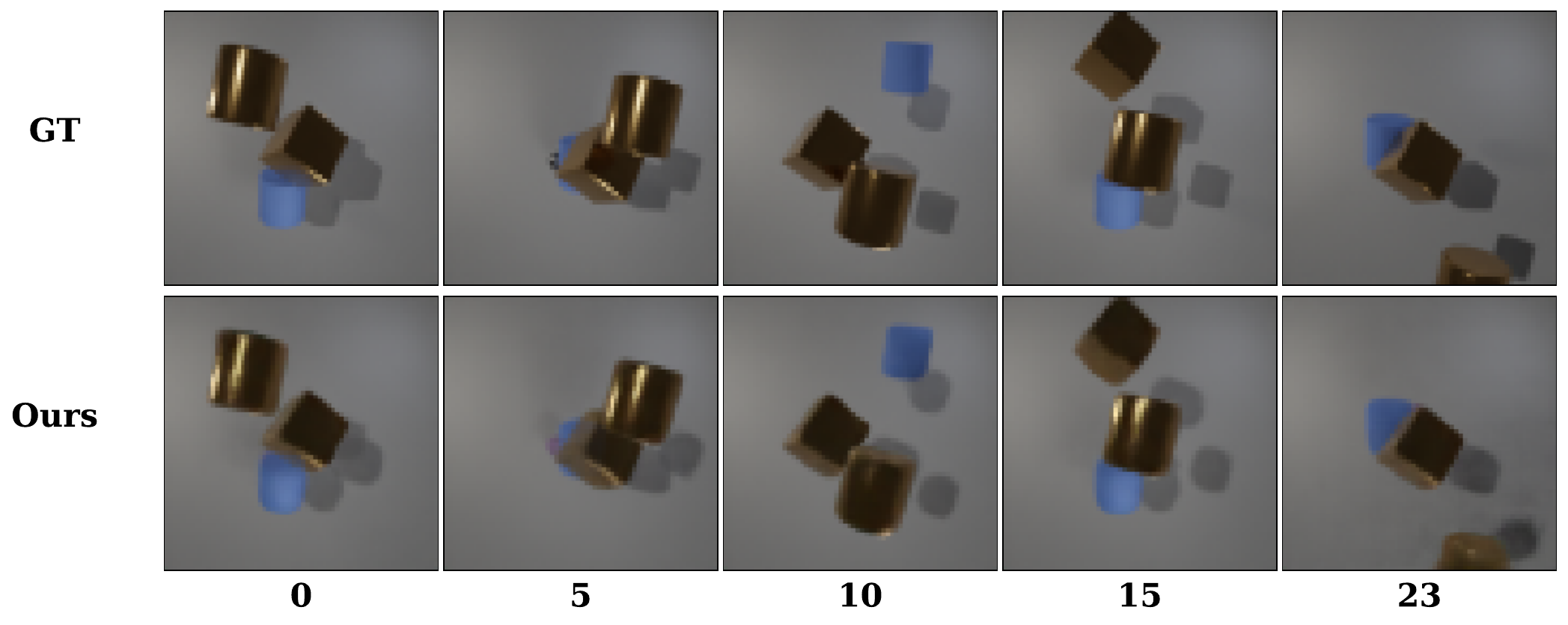}
    \end{minipage}
    \caption{Qualitative performance of our model for the dataset variant B: Increased gravitational constant}
\end{figure}

\begin{figure}[H]
\centering
    \begin{minipage}{.46\textwidth}
    \includegraphics[width=\columnwidth]{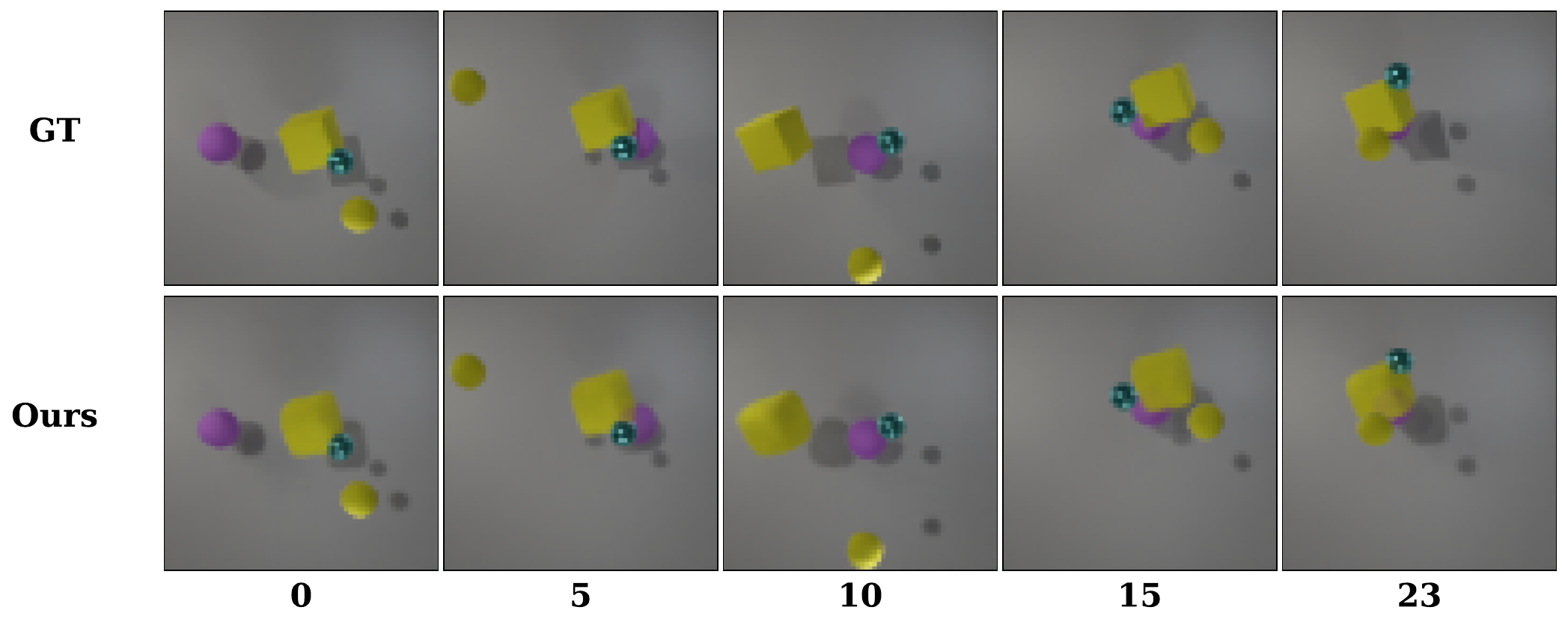}
    \end{minipage}%
    \begin{minipage}{.46\textwidth}
    \includegraphics[width=\columnwidth]{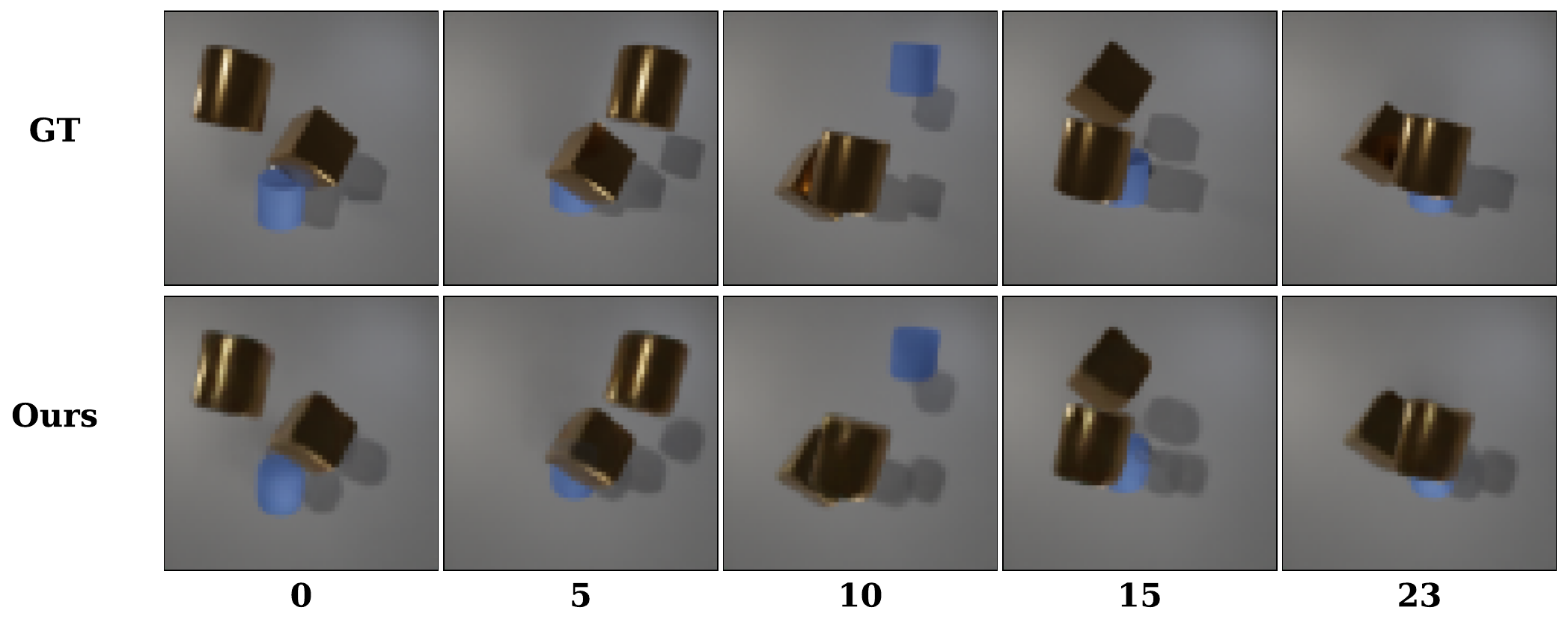}
    \end{minipage}
    \caption{Qualitative performance of our model for the dataset variant C: Tripled force}
\end{figure}

\begin{figure}[H]
\centering
    \begin{minipage}{.46\textwidth}
    \includegraphics[width=\columnwidth]{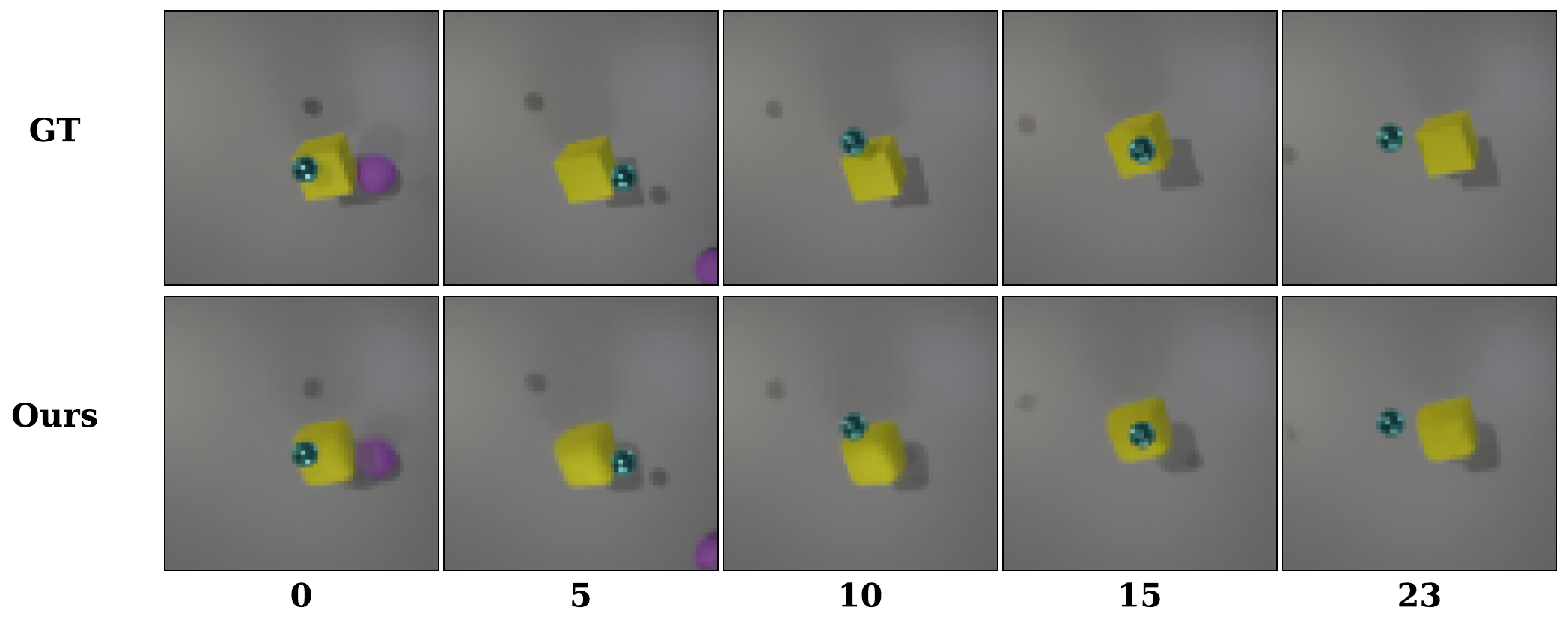}
    \end{minipage}%
    \begin{minipage}{.46\textwidth}
    \includegraphics[width=\columnwidth]{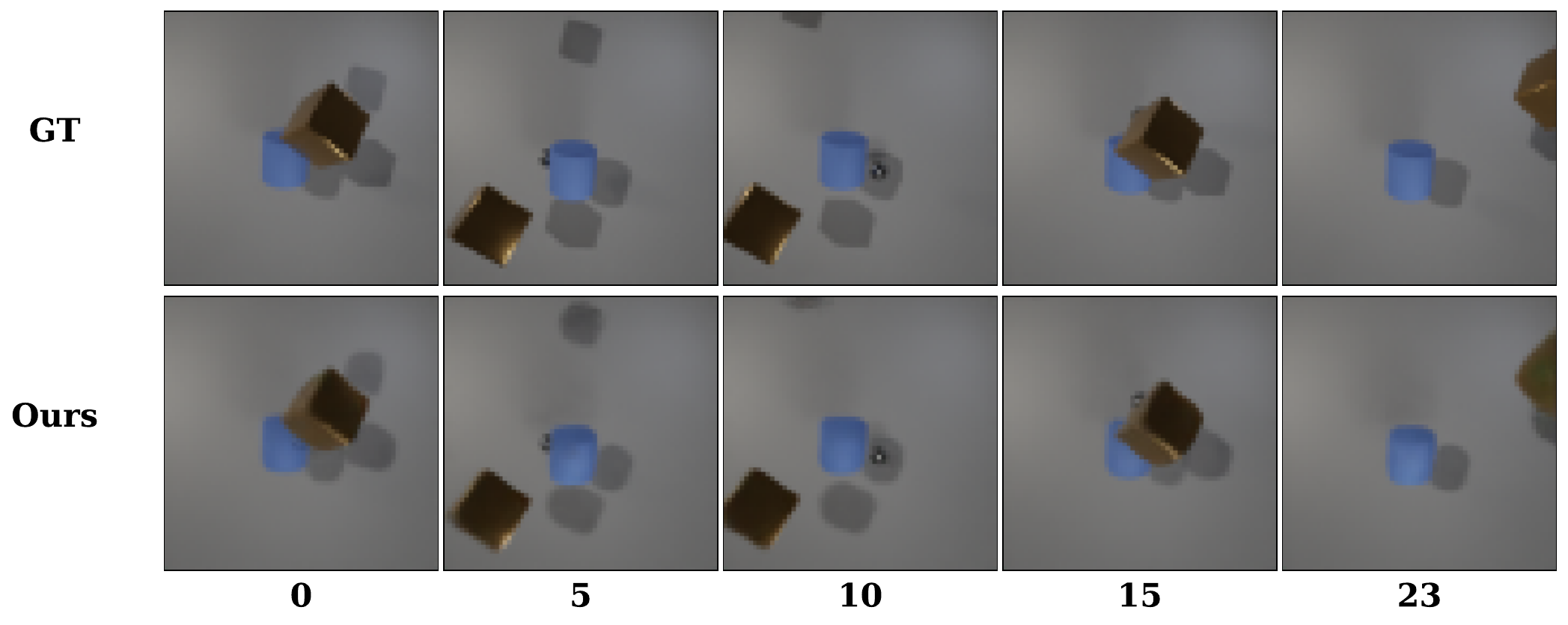}
    \end{minipage}
    \caption{Qualitative performance of our model for the dataset variant D: Objects are repulsed}
\end{figure}

\begin{figure}[H]
\centering
    \begin{minipage}{.46\textwidth}
    \includegraphics[width=\columnwidth]{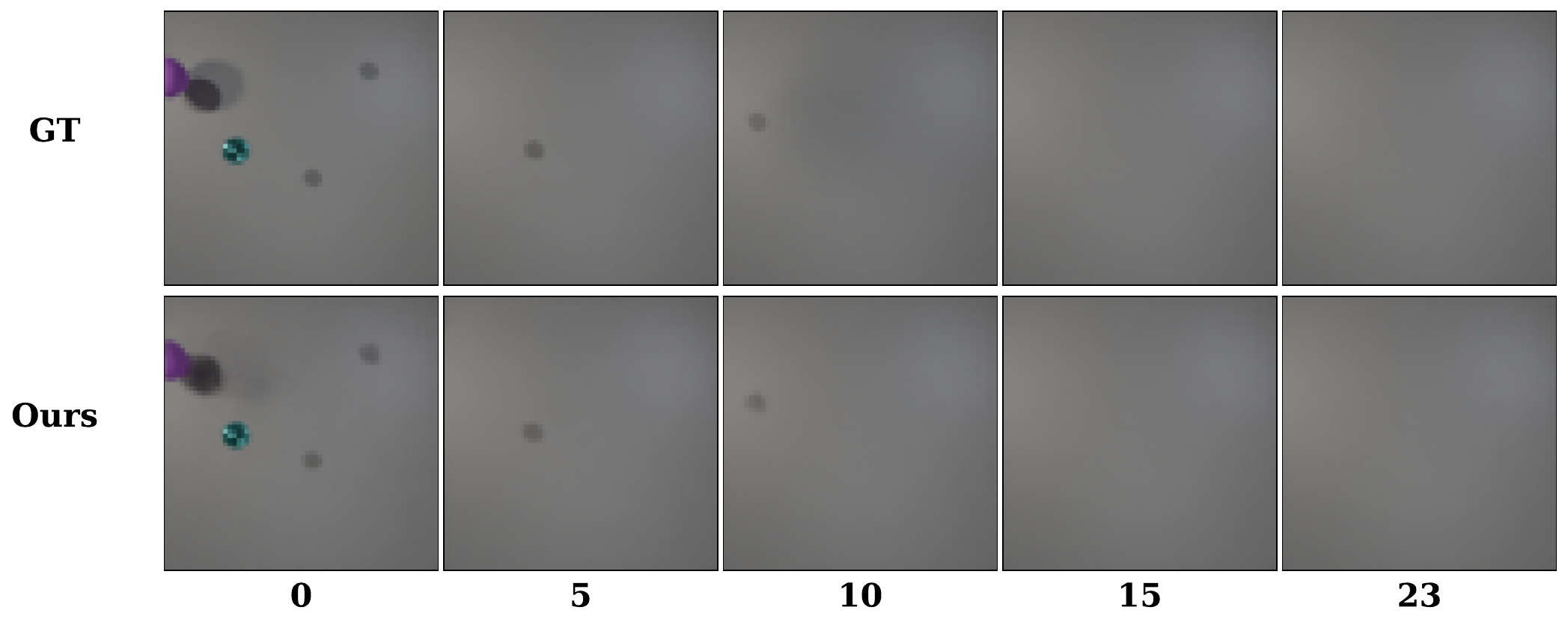}
    \end{minipage}%
    \begin{minipage}{.46\textwidth}
    \includegraphics[width=\columnwidth]{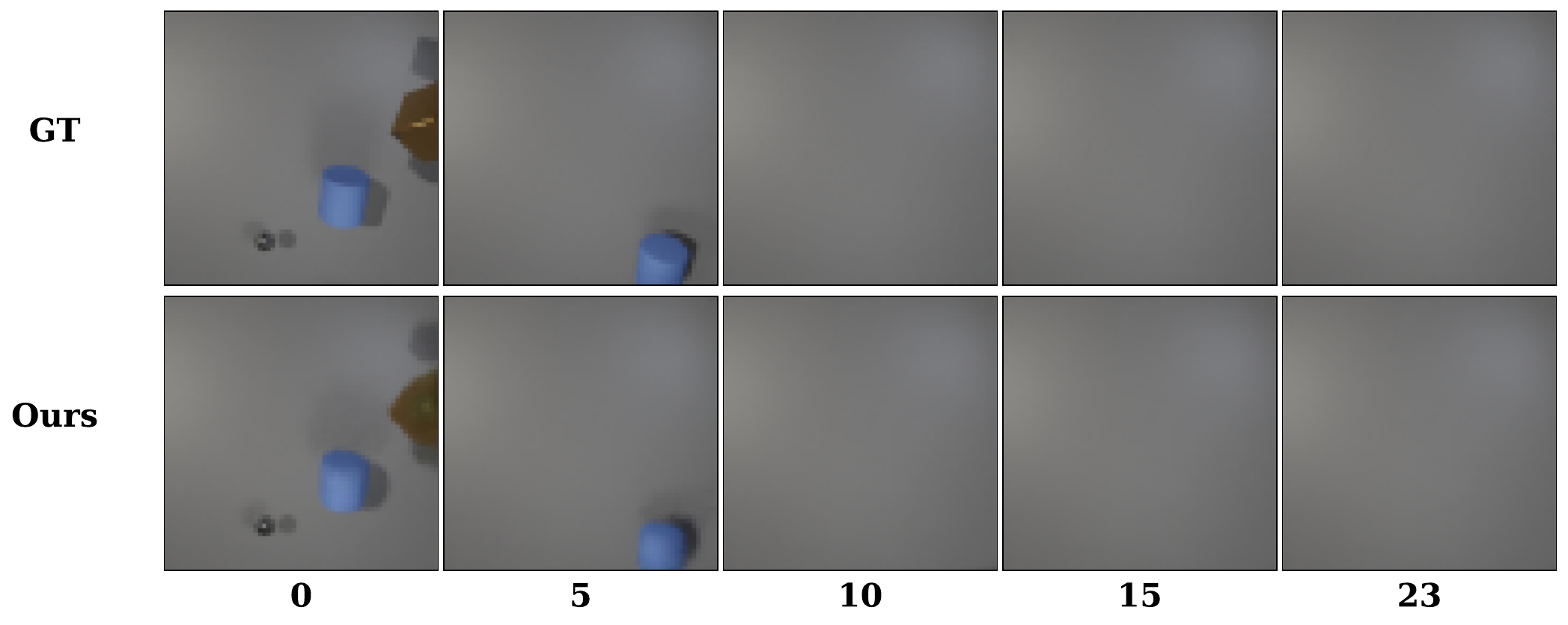}
    \end{minipage}
    \caption{Qualitative performance of our model for the dataset variant E: No forces}
\end{figure}

\begin{figure}[H]
\centering
    \begin{minipage}{.46\textwidth}
    \includegraphics[width=\columnwidth]{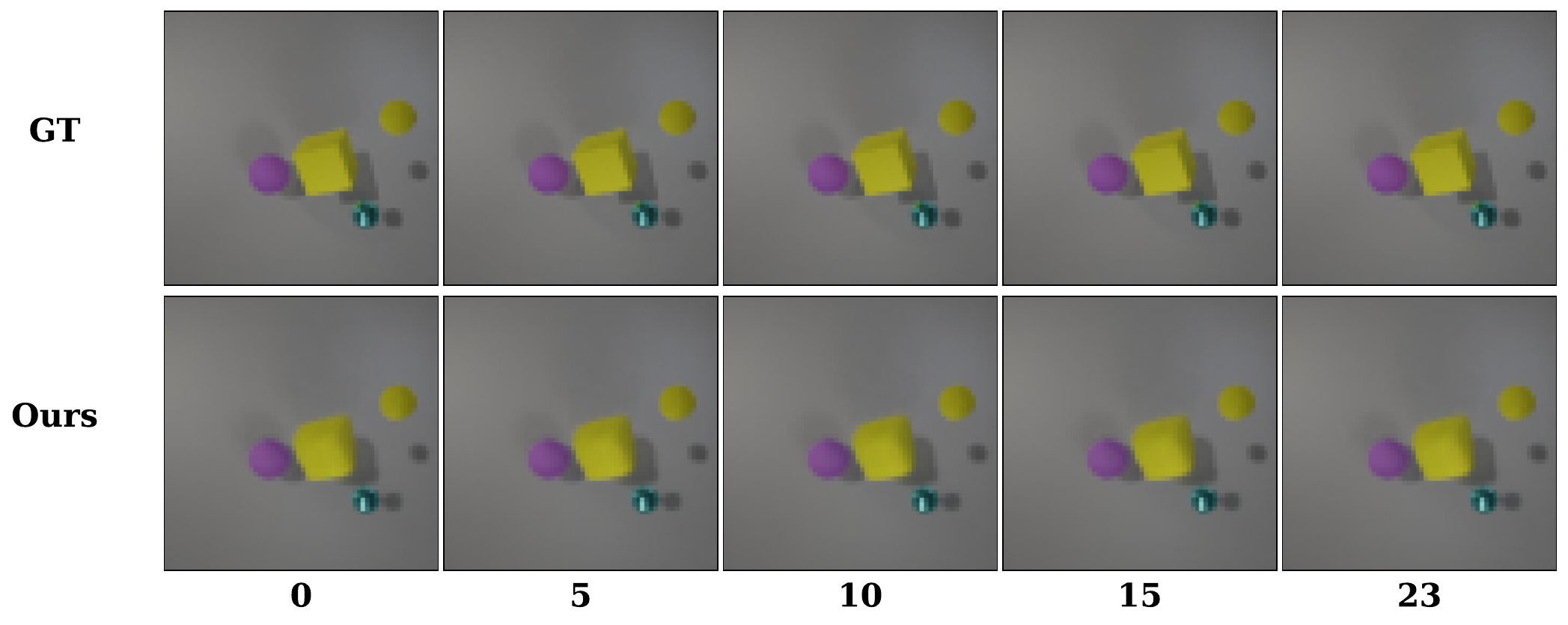}
    \end{minipage}%
    \begin{minipage}{.46\textwidth}
    \includegraphics[width=\columnwidth]{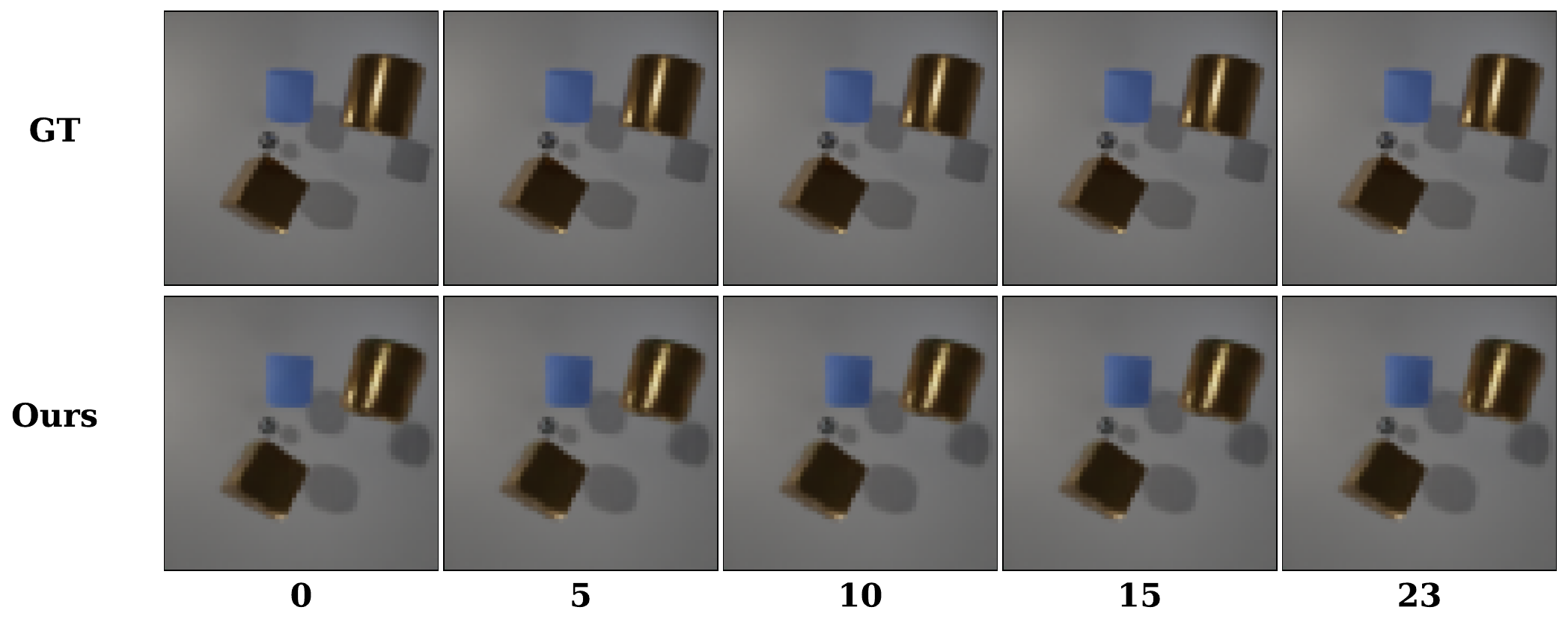}
    \end{minipage}
    \caption{Qualitative performance of our model for the dataset variant F: No forces, no velocities}
\end{figure}

\begin{figure}[H]
\centering
    \begin{minipage}{.46\textwidth}
    \includegraphics[width=\columnwidth]{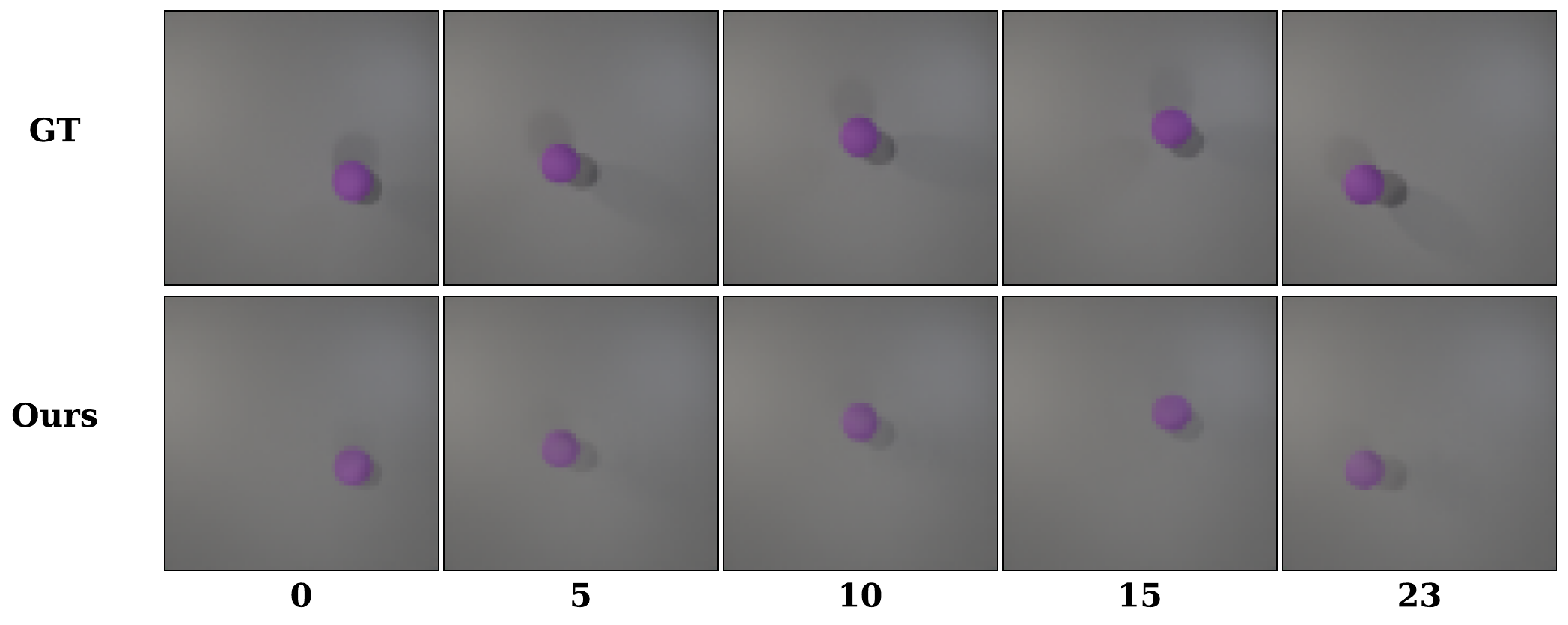}
    \end{minipage}%
    \begin{minipage}{.46\textwidth}
    \includegraphics[width=\columnwidth]{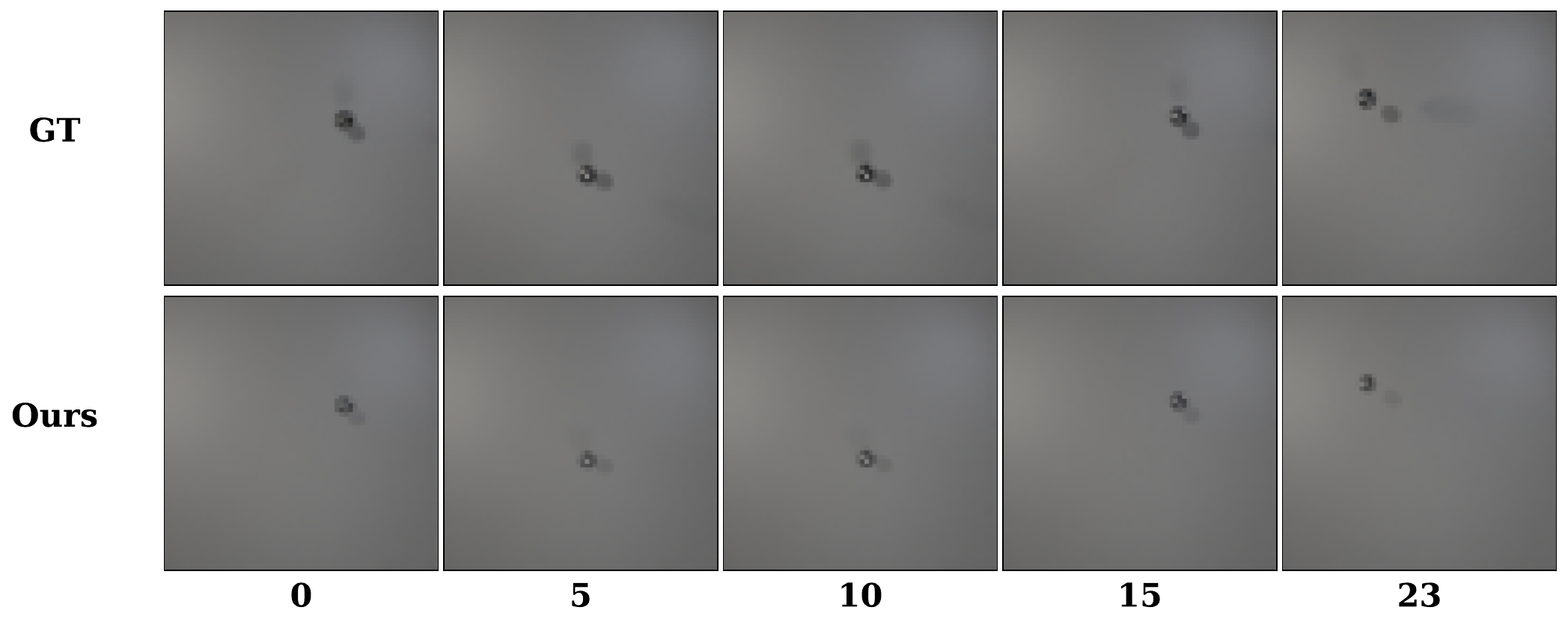}
    \end{minipage}
    \caption{Qualitative performance of our model for the dataset variant G: Trajectory following}
\end{figure}

\begin{figure}[H]
\centering
    \begin{minipage}{.46\textwidth}
    \includegraphics[width=\columnwidth]{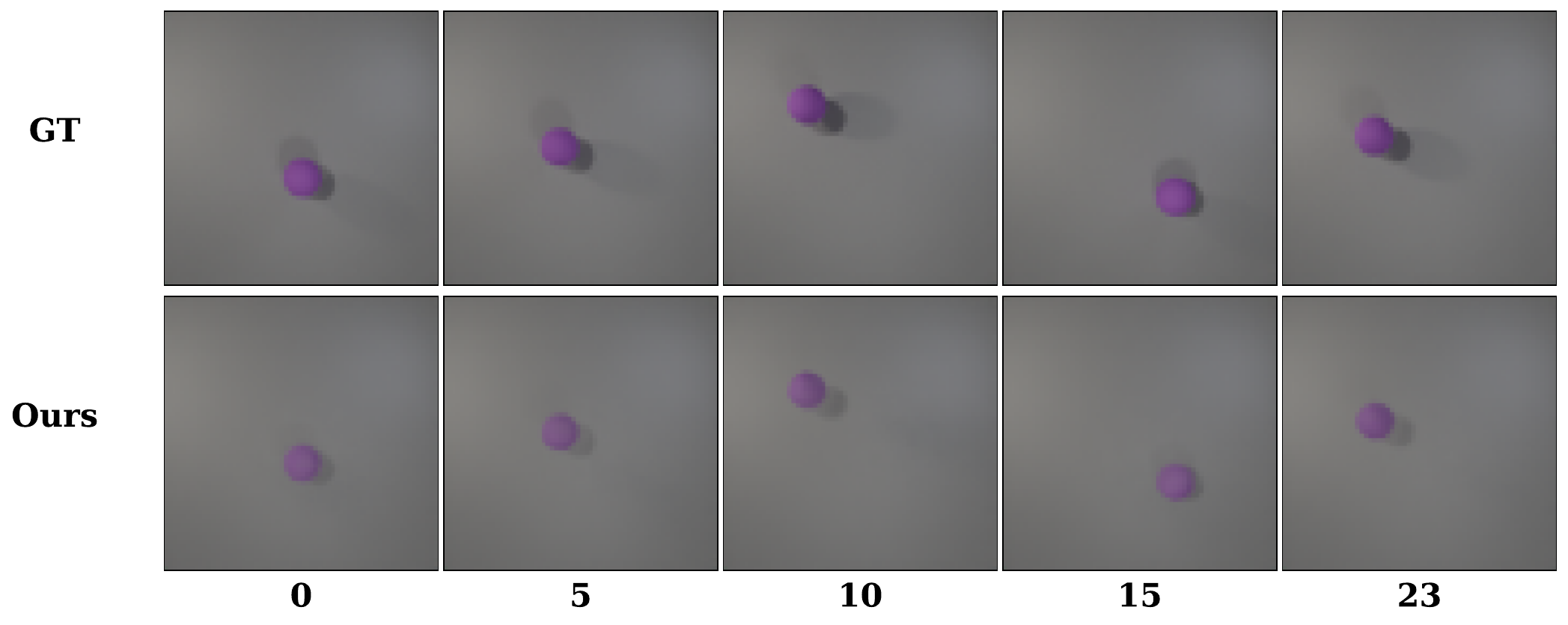}
    \end{minipage}%
    \begin{minipage}{.46\textwidth}
    \includegraphics[width=\columnwidth]{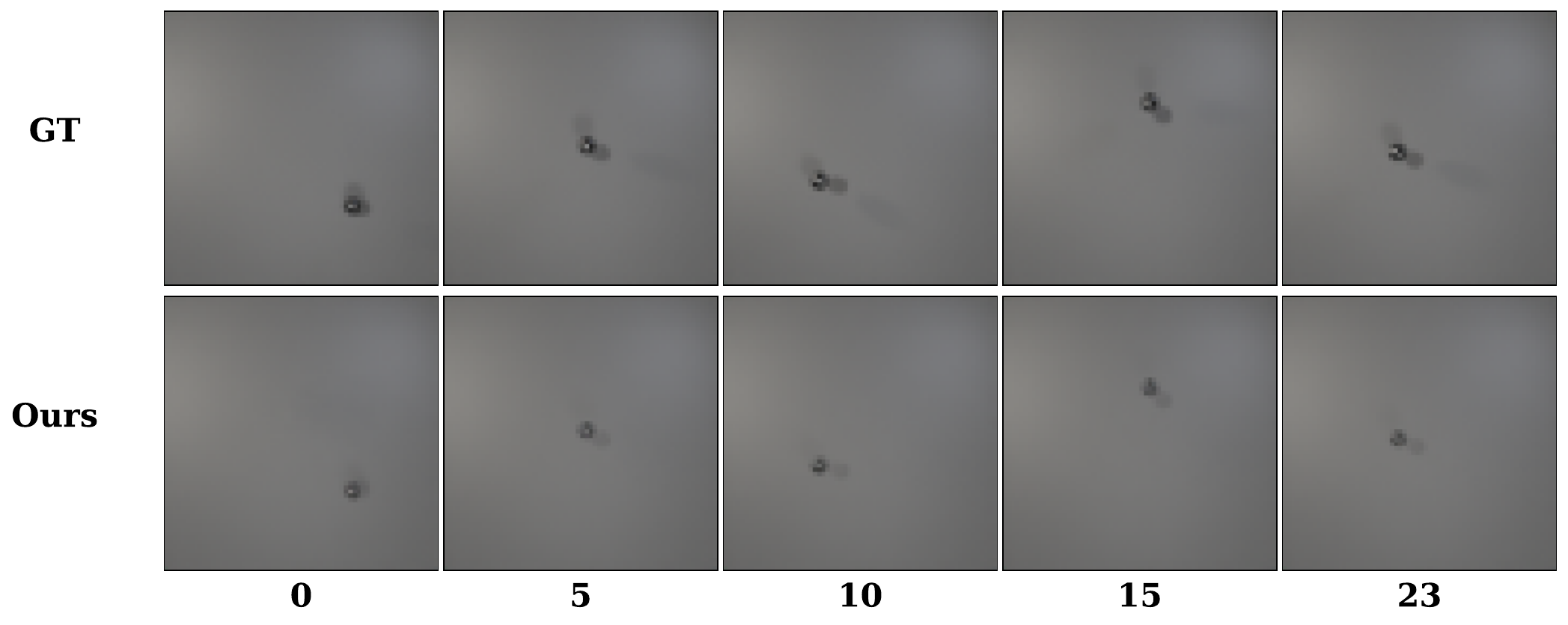}
    \end{minipage}
    \caption{Qualitative performance of our model for the dataset variant H: Random location}
\end{figure}

\newpage

\section{Integrated Function Details}\label{apd:f}
This section shows the functions integrated in our model. All functions first calculate the appropriate acceleration $a$ before applying it in a semi-implicit euler integration step with a step size of $\Delta t$.

For the Orbits dataset each objects state consists of position $p$ and velocity $v$. The environmental constants correspond to the gravitational constant $g$ and object mass $m$. Given $N$ objects in the scene at video frame $t$, the object state of the next time step $t+1$ for any object $n$ is obtained as follows:

\begin{align}
    F_{t,n} &= \sum_{\substack{i=0\\i\neq n}}^N \frac{(p_{t,i} - p_{t,n})}{|(p_{t,i} - p_{t,n})|}\frac{gm}{|(p_{t,i} - p_{t,n})|^2} \\
    a_{t,n} &= \frac{F_{t,n}}{m}\\
    v_{t+1,n} &= v_{t,n} + \Delta t a_{t,n}\\
    p_{t+1,n} &= p_{t,n} + \Delta t v_{t+1,n}
\end{align}

For the Acrobot dataset the per-frame state consists of the pendulum angles $\theta_1$ and $\theta_2$ and their angular velocities $\dot{\theta}_1$ and $\dot{\theta}_2$. The environmental constants consist of the pendulum masses $m_1$ and $m_2$, the pendulum lengths $l_1$ and $l_2$, the link center of mass $c_1$ and $c_2$, the inertias $I_1$ and $I_2$, and the gravitational constant $G$. The pendulum state of the next time step $t+1$ is calculated as follows:

\begin{align}
    \delta_{1_t} &= m_1 c_1^2 + m_2 (l_1^2 + c_2^2 + 2l_1 c_2 \cos{(\theta_{2_t})}) + I_1 + I_2\\
    \delta_{2_t} &= m_2 (c_2^2 + l_1 c_2 \cos{(\theta_{2_t})}) + I_2\\
    \phi_{2_t} &= m_2 c_2 G \cos{(\theta_{1_t} + \theta_{2_t} - \frac{\pi}{2})}\\
    \phi_{1_t} &= -m_2 l_1 c_2 \dot{\theta}_{2_t}^2 \sin{(\theta_{2_t})} - 2 m_2 l_1 c_2 \dot{\theta}_{2_t} \dot{\theta}_{1_t} \sin{(\theta_{2_t})}\\
    &\;\;\;\; + (m_1 c_1 + m_2 l_1) G \cos{(\theta_{1_t} - \frac{\pi}{2})} + \phi_{2_t}\\
    \ddot{\theta}_{2_t} &= \frac{\frac{\delta_{2_t}}{\delta_{1_t}} \phi_{1,t} - m_2 l_1 c_2 \dot{\theta}_{1,t}^2 \sin{(\theta_{2,t})} - \phi_{2,t}}{m_2 c_2^2 + I_2 - \frac{\delta_{2_t}^2}{\delta_{1_t}}}\\
    \ddot{\theta}_{1_t} &= -\frac{\delta_2 \ddot{\theta}_{2_t} + \phi_{1_t}}{\delta_{1_t}}\\
    \dot{\theta}_{1_{t+1}} &= \dot{\theta}_{1_{t}} + \Delta t \ddot{\theta}_{1_t}\\
    \dot{\theta}_{2_{t+1}} &= \dot{\theta}_{2_{t}} + \Delta t \ddot{\theta}_{2_t}\\
    \theta_{1_{t+1}} &= \theta_{1_{t}} + \Delta t \dot{\theta}_{1_{t+1}}\\
    \theta_{2_{t+1}} &= \theta_{2_{t}} + \Delta t \dot{\theta}_{2_{t+1}}
\end{align}

The Pendulum Camera dataset follows the same equations of the Acrobot dataset to obtain an updated pendulum state. Afterwards, this state is used to obtain the new camera position $p_{c_{t+1}}$:

\begin{align}
p_{c_{t+1}} = 
\begin{bmatrix}
p_x\\
p_y\\
p_z\\
\end{bmatrix}
&=
\begin{bmatrix}
-2l_1 \sin{(\theta_{1_{t+1}})} - l_2 \sin{(\theta_{1_{t+1}} + \theta_{2_{t+1}})} \\
2l_1 \cos{(\theta_{1_{t+1}})} + l_2 \cos{(\theta_{1_{t+1}} + \theta_{2_{t+1}})}\\
10
\end{bmatrix}
\end{align}

\section{\gls*{mpc} Details}\label{apd:mpc}
We set the control objective as the maximization of the potential energy---i.e. both pendulums oriented upwards---and the minimization of the kinetic energy---i.e. resting pendulums. The system model corresponds to our integrated function $F$ and due to already being discretized does not require further processing. We use a controller with a prediction horizon of 150 steps and store the predicted torque action sequence for the next 75 frames.

\newpage

\subsection{Qualitative Results}
This section shows additional qualitative results for the \gls*{mpc} task.

\begin{figure}[h]
\centering
    \includegraphics[width=.85\columnwidth]{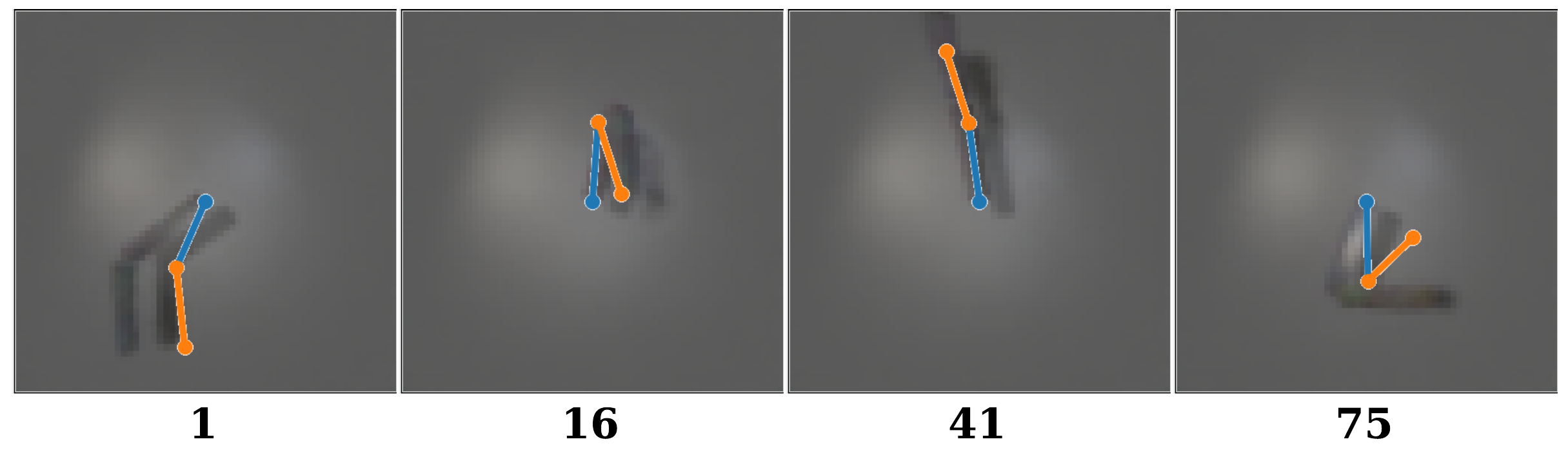}
     \includegraphics[width=.85\columnwidth]{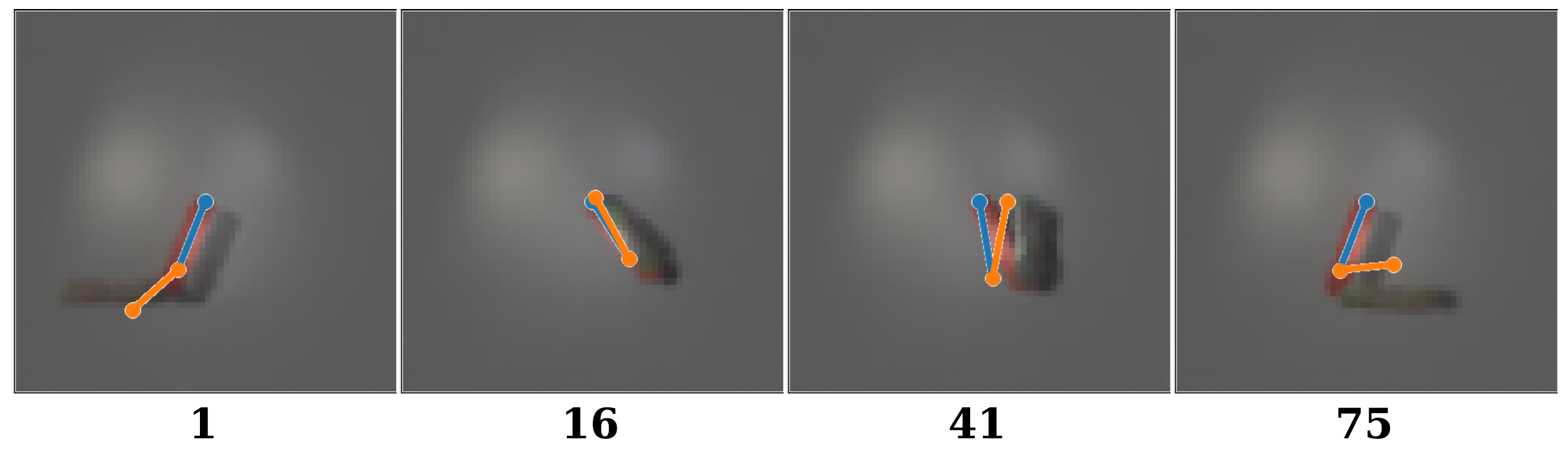}
     \includegraphics[width=.85\columnwidth]{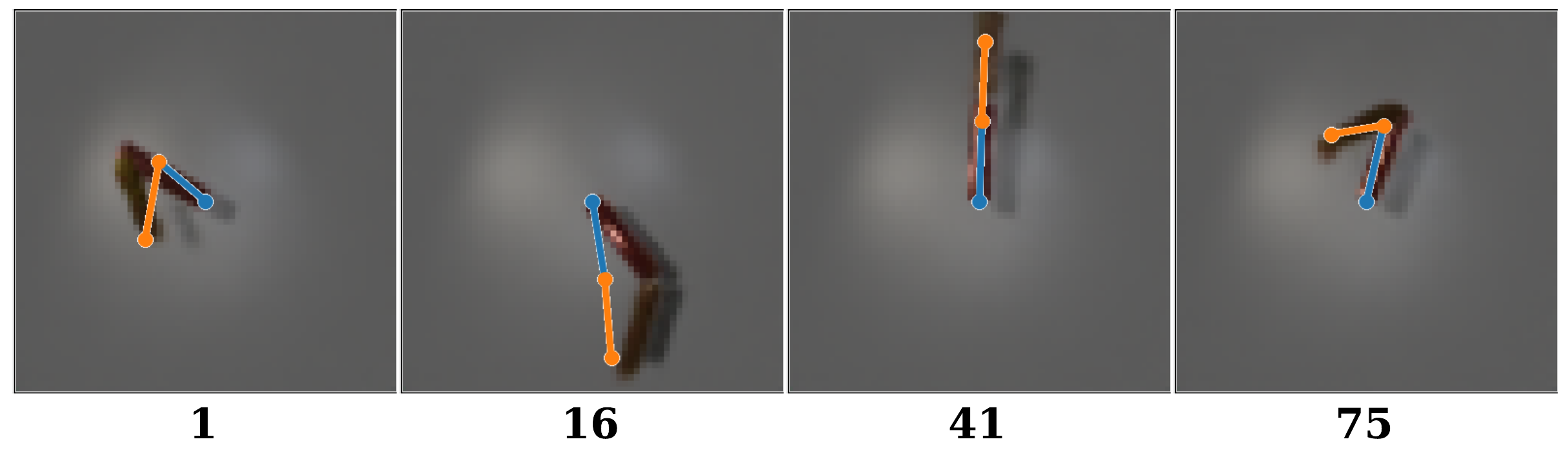}
     \includegraphics[width=.85\columnwidth]{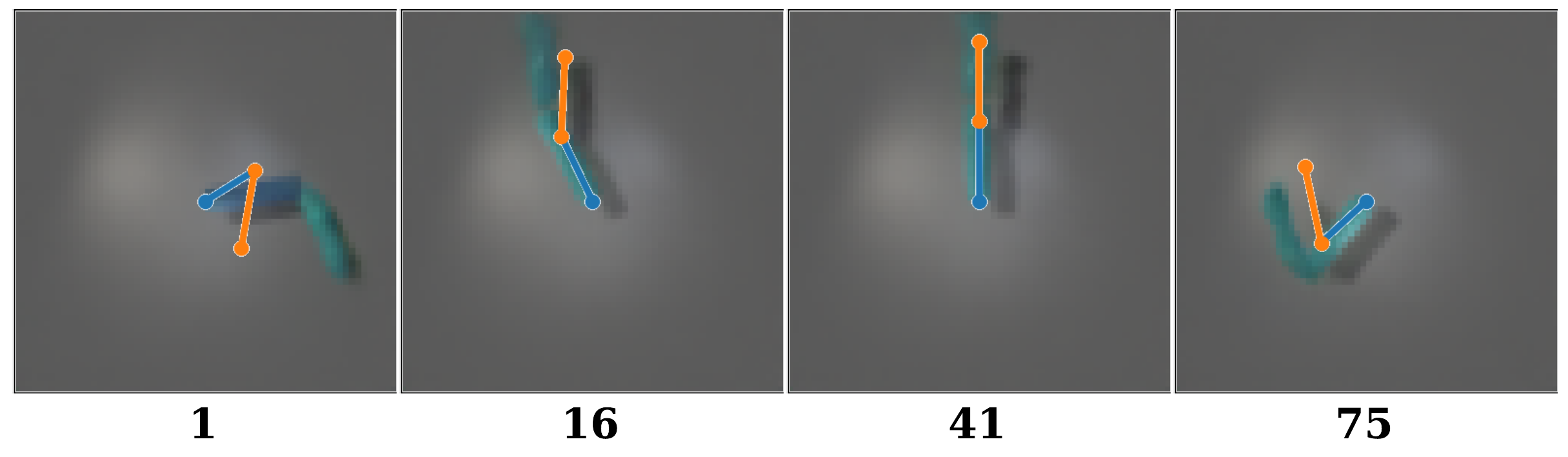}
    \vspace{-0.4cm}
    \caption{Qualitative results for the \gls*{mpc} task.}
\end{figure}

\end{document}